\documentclass{article}

\usepackage{microtype}
\usepackage{graphicx}

\usepackage{booktabs} 
\usepackage{wrapfig,lipsum}
\usepackage{subcaption}

\usepackage{url}
\usepackage{hyperref}
\usepackage{inconsolata}

\newcommand{\scrX}{\mathcal{X}}
\newcommand{\scrY}{\mathcal{Y}}

\usepackage{tabu}
\usepackage{amsmath,amsthm,verbatim,amssymb,amsfonts,amscd,mathrsfs}
\usepackage{cleveref}

\theoremstyle{plain}
\newtheorem{theorem}{Theorem}

\newtheorem{lemma}{Lemma}

\theoremstyle{remark}

\theoremstyle{definition}

\newcommand{\todo}[1]{}
\renewcommand{\todo}[1]{{\color{red} TODO: {#1}}}

\newcommand{\grad}{\nabla}

\newcommand{\opt}{^*}

\def\1#1{1\left[ #1 \right]} 

\def\norm#1{\left\| #1 \right\|}
\def\paren#1{\left( #1 \right)}     

\usepackage{amsmath,amsfonts,bm}

\def\eqref#1{equation~\ref{#1}}

\def\1{\bm{1}}

\def\eps{{\epsilon}}

\def\vtheta{{\bm{\theta}}}

\def\vg{{\bm{g}}}

\def\vv{{\bm{v}}}
\def\vw{{\bm{w}}}

\def\mA{{\bm{A}}}

\def\mU{{\bm{U}}}
\def\mV{{\bm{V}}}

\DeclareMathAlphabet{\mathsfit}{\encodingdefault}{\sfdefault}{m}{sl}
\SetMathAlphabet{\mathsfit}{bold}{\encodingdefault}{\sfdefault}{bx}{n}

\newcommand{\E}{\mathbb{E}}

\newcommand{\R}{\mathbb{R}}

\DeclareMathOperator{\sign}{sign}
\DeclareMathOperator{\Tr}{Tr}

\usepackage[accepted]{icml2020}

\icmltitlerunning{Test-Time Training with Self-Supervision
for Generalization under Distribution Shifts}

\begin{document}

\twocolumn[
\icmltitle{Test-Time Training with Self-Supervision\\
	for Generalization under Distribution Shifts}



\icmlsetsymbol{equal}{*}

\begin{icmlauthorlist}
\icmlauthor{Yu Sun}{to}
\icmlauthor{Xiaolong Wang}{to,ucsd}
\icmlauthor{Zhuang Liu}{to}
\icmlauthor{John Miller}{to}
\icmlauthor{Alexei A. Efros}{to}
\icmlauthor{Moritz Hardt}{to}
\end{icmlauthorlist}

\icmlaffiliation{to}{University of California, Berkeley}
\icmlaffiliation{ucsd}{University of California, San Diego}

\icmlcorrespondingauthor{Yu Sun}{yusun@berkeley.edu}

\icmlkeywords{Test-Time Training, distribution shifts}

\vskip 0.3in
]



\printAffiliationsAndNotice{} 

\begin{abstract}
In this paper, we propose Test-Time Training, a general approach for improving the performance of predictive models when training and test data come from different distributions. We turn a single unlabeled test sample into a self-supervised learning problem, on which we update the model parameters before making a prediction. This also extends naturally to data in an online stream. Our simple approach leads to improvements on diverse image classification benchmarks aimed at evaluating robustness to distribution shifts.
\end{abstract}

\section{Introduction}
\label{intro}

Supervised learning remains notoriously weak at generalization under distribution shifts.
Unless training and test data are drawn from the same distribution, even seemingly minor differences turn out to defeat state-of-the-art models \citep{recht2018cifar}.
Adversarial robustness and domain adaptation are but a few existing paradigms that try to \textit{anticipate} differences between the training and test distribution with either topological structure or data from the test distribution available during training.
We explore a new take on generalization that \textit{does not} anticipate the distribution shifts, but instead learns from them at test time.

We start from a simple observation. 
The unlabeled test sample $x$ presented at test time gives us a hint about the distribution from which it was drawn. 
We propose to take advantage of this hint on the test distribution by allowing the model parameters $\vtheta$
to depend on the test sample $x$, but not its unknown label $y$.
The concept of a variable decision boundary $\vtheta(x)$ is powerful in theory since it breaks away from the limitation of fixed model capacity (see additional discussion in \Cref{vdb}), but the design of a feedback mechanism from $x$ to $\vtheta(x)$ raises new challenges in practice that we only begin to address here.

Our proposed test-time training method creates a self-supervised learning problem based on this single test sample $x$, updating $\vtheta$ at test time before making a prediction. 
Self-supervised learning uses an auxiliary task that automatically creates labels from unlabeled inputs.
In our experiments, we use the task of rotating each input image by a multiple of 90 degrees and predicting its angle~\citep{gidaris2018unsupervised}.

This approach can also be easily modified to work outside the standard supervised learning setting. 
If several test samples arrive in a batch, we can use the entire batch for test-time training.
If samples arrive in an online stream, we obtain further improvements by keeping the state of the parameters.
After all, prediction is rarely a single event. The online version can be the natural mode of deployment under the additional assumption that test samples are produced by the same or smoothly changing distribution shifts.

We experimentally validate our method in the context of object recognition on several standard benchmarks. These include images with diverse types of corruption at various levels \citep{hendrycks2019benchmarking},  video frames of moving objects \citep{shankar2019systematic}, and a new test set of unknown shifts collected by \cite{recht2018cifar}. 
Our algorithm makes substantial improvements under distribution shifts, while maintaining the same performance on the original distribution.

In our experiments, we compare with a strong baseline (labeled joint training) that uses both supervised and self-supervised learning at training-time, but keeps the model fixed at test time.
Recent work shows that \emph{training-time} self-supervision improves robustness \citep{hendrycks2019using}; our joint training baseline corresponds to an improved implementation of this work. 
A comprehensive review of related work follows in \Cref{related}.

We complement the empirical results with theoretical investigations in \Cref{theory}, and establish an intuitive sufficient condition on a convex model of when Test-Time Training helps; this condition, roughly speaking, is to have correlated gradients between the loss functions of the two tasks.
\newcommand\blfootnote[1]{%
  \begingroup
  \renewcommand\thefootnote{}\footnote{#1}%
  \addtocounter{footnote}{-1}%
  \endgroup
}
\blfootnote{Project website: {\scriptsize \url{https://test-time-training.github.io/}}.}
\newpage
\section{Method}
\label{method}
This section describes the algorithmic details of our method.
To set up notation, consider a standard $K$-layer neural network with parameters~$\theta_k$ for layer~$k$. The stacked parameter vector $\vtheta=(\theta_1,\dots,\theta_K)$ specifies the entire model for a classification task with loss function~$l_m(x, y; \vtheta)$ on the test sample $(x,y)$. We call this the \emph{main task}, as indicated by the subscript of the loss function. 

We assume to have training data $(x_1,y_1),\dots,(x_n, y_n)$ drawn i.i.d.~from a distribution~$P$. Standard empirical risk minimization solves the optimization problem:
\begin{equation} \label{optimize}
\min_\vtheta \frac{1}{n}\sum_{i=1}^{n} l_m(x_i, y_i; \vtheta).
\end{equation}
Our method requires a \emph{self-supervised auxiliary task} with loss function~$l_s(x)$. 
In this paper, we choose the rotation prediction task \citep{gidaris2018unsupervised}, which has been demonstrated to be simple and effective at feature learning for convolutional neural networks.
The task simply rotates $x$ in the image plane by one of 0, 90, 180 and 270 degrees and have the model predict the angle of rotation as a four-way classification problem. 
Other self-supervised tasks in \Cref{related} might also be used for our method.

The auxiliary task shares some of the model parameters
$\vtheta_e=(\theta_1,\dots,\theta_\kappa)$ up to a
certain~$\kappa\in\{1,\dots,K\}.$
We designate those $\kappa$ layers as a
\emph{shared feature extractor}. 
The auxiliary task uses its own task-specific parameters 
$\vtheta_s = (\theta'_{\kappa+1},\dots, \theta'_{K})$. 
We call the unshared parameters $\vtheta_s$ the \emph{self-supervised task branch}, and $\vtheta_m=(\theta_{\kappa+1},\dots,\theta_K)$ the
\emph{main task branch}.
Pictorially, the joint architecture is a $Y$-structure with a shared bottom and two branches.
For our experiments, the self-supervised task branch has the same architecture as the main branch, except for the output dimensionality of the last layer due to the different number of classes in the two tasks.

Training is done in the fashion of multi-task learning \citep{caruana1997multitask}; the model is trained on both tasks on the same data drawn from~$P$. 
Losses for both tasks are added together, and gradients are taken for the collection of all parameters. The joint training problem is therefore
\begin{equation}
\label{optimize_train}
\min_{\vtheta_e,\vtheta_m,\vtheta_s}
\frac{1}{n}\sum_{i=1}^{n} 
l_m(x_i, y_i; \vtheta_m, \vtheta_e)
+ l_s(x_i; \vtheta_s, \vtheta_e).
\end{equation}

Now we describe the standard version of Test-Time Training on a single test sample $x$.
Simply put, Test-Time Training fine-tunes the shared feature extractor $\vtheta_e$ by minimizing the auxiliary task loss on $x$. 
This can be formulated as
\begin{equation}
    \label{optimize_test}
    \min_{\vtheta_e}l_s(x; \vtheta_s, \vtheta_e).
\end{equation}
Denote $\vtheta_e\opt$ the (approximate) minimizer of \autoref{optimize_test}.
The model then makes a prediction using the updated parameters 
$\vtheta(x) = (\vtheta_e\opt, \vtheta_m)$.
Empirically, the difference is negligible between minimizing \autoref{optimize_test} over $\vtheta_e$ versus over both $\vtheta_e$ and $\vtheta_s$.
Theoretically, the difference exists only when optimization is done with more than one gradient step.

Test-Time Training naturally benefits from standard data augmentation techniques. On each test sample $x$, we perform the exact same set of random transformations as for data augmentation during training, to form a batch only containing these augmented copies of $x$ for Test-Time Training.

\paragraph{Online Test-Time Training.}
In the standard version of our method, the optimization problem in
\autoref{optimize_test} is always initialized with parameters 
$\vtheta = (\vtheta_e,\vtheta_s)$ 
obtained by minimizing \autoref{optimize_train}.
After making a prediction on $x$, $\vtheta_e\opt$ is discarded.
Outside of the standard supervised learning setting, 
when the test samples arrive online sequentially, 
the online version solves the same optimization problem as in \autoref{optimize_test} to update the shared feature extractor $\vtheta_e$.
However, on test sample $x_t$, 
$\vtheta$ is instead initialized with $\vtheta(x_{t-1})$ updated on the previous sample $x_{t-1}$.
This allows $\vtheta(x_t)$ to take advantage of the distributional information available in $x_1, \dots,x_{t-1}$ as well as $x_t$.

\section{Empirical Results}
\label{results}
We experiment with both versions of our method (standard and online) on three kinds of benchmarks for distribution shifts, presented here in the order of visually low to high-level.
Our code is available at the project website.

\begin{figure*}
	\centering
	\includegraphics[width=1.0\textwidth]{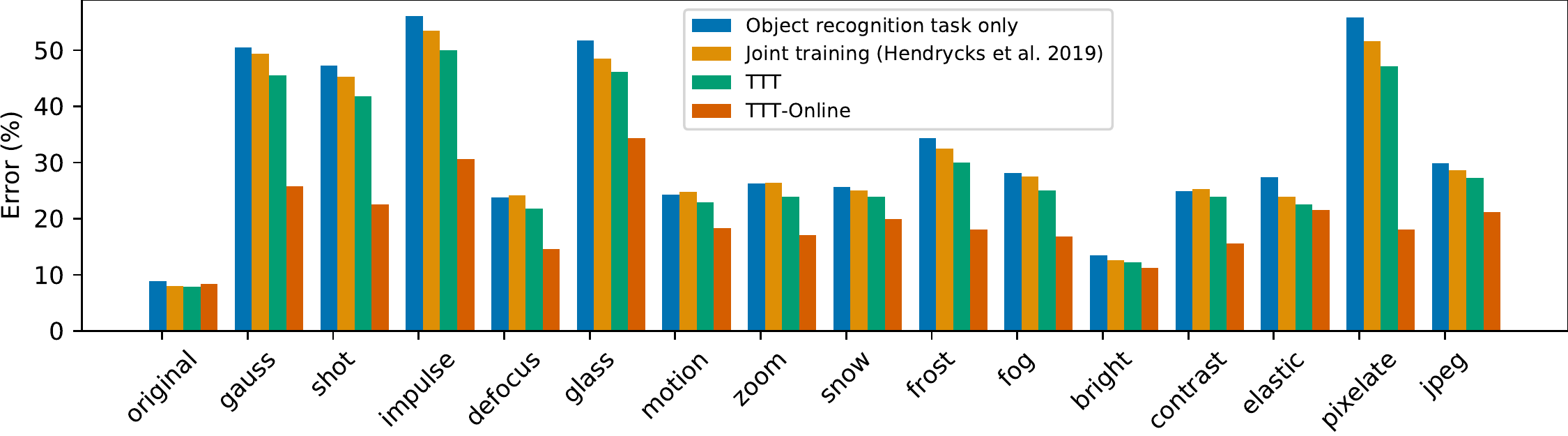}
	\caption{
	{\bf Test error (\%) on CIFAR-10-C with level 5 corruptions.} We compare our approaches, Test-Time Training (TTT) and its online version (TTT-Online), with two baselines: object recognition without self-supervision, and joint training with self-supervision but keeping the model fixed at test time. 
	TTT improves over the baselines and TTT-Online improves even further. 
	}
	\label{fig:cc}
\end{figure*}
\paragraph{Network details.}
Our architecture and hyper-parameters are consistent across all experiments.  We use ResNets \citep{he2016identity}, which are constructed differently for CIFAR-10 \citep{krizhevsky2009learning} (26-layer) and ImageNet \citep{ILSVRC15} (18-layer). The CIFAR-10 dataset contains 50K images for training, and 10K images for testing. The ImageNet contains 1.2M images for training and the 50K validation images are used as the test set. ResNets on CIFAR-10 have three groups, each containing convolutional layers with the same number of channels and size of feature maps; our splitting point is the end of the second group. ResNets on ImageNet have four groups;  our splitting point is the end of the third group. 

We use Group Normalization (GN) instead of Batch Normalization (BN) in our architecture, since BN has been shown to be ineffective when training with small batches, for which the estimated batch statistics are not accurate \citep{ioffe2015batch}. This technicality hurts Test-Time Training since each batch only contains (augmented) copies of a single image. Different from BN, GN is not dependent on batch size and achieves similar results on our baselines. We report results with BN in \Cref{results_additional} of the appendix for completeness. 
We directly compare our architecture to that of \citet{hendrycks2018using} in \autoref{reviewer_stuff}.

\begin{figure*}[t]
	\centering
	\includegraphics[width=1.0\textwidth]{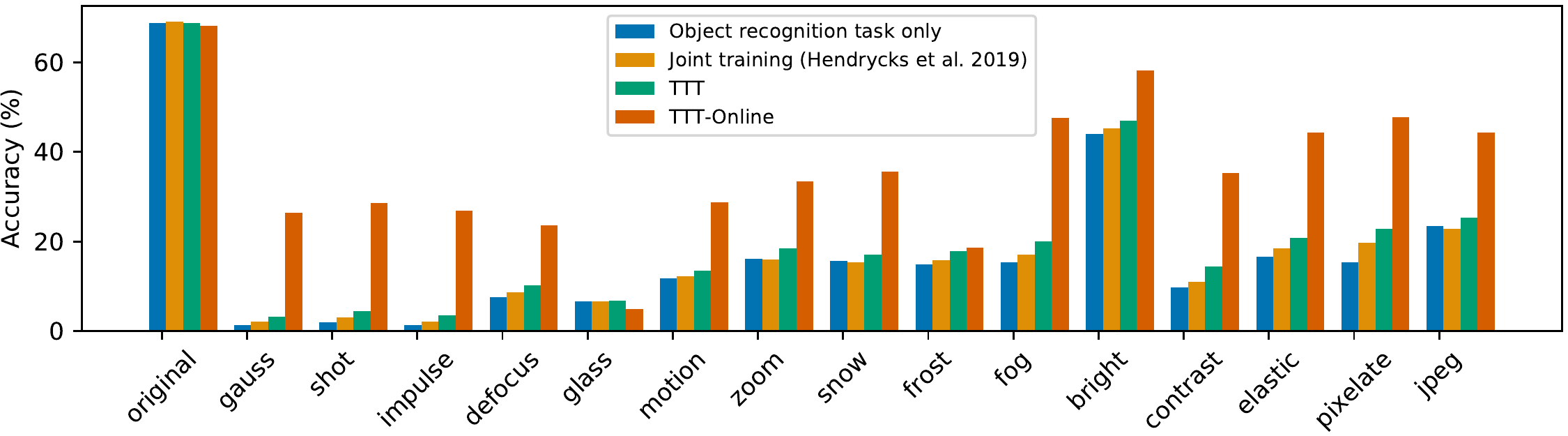}
		\vspace{-2ex}
	\begin{subfigure}[b]{1.0\textwidth}
		\includegraphics[width=0.325\textwidth]{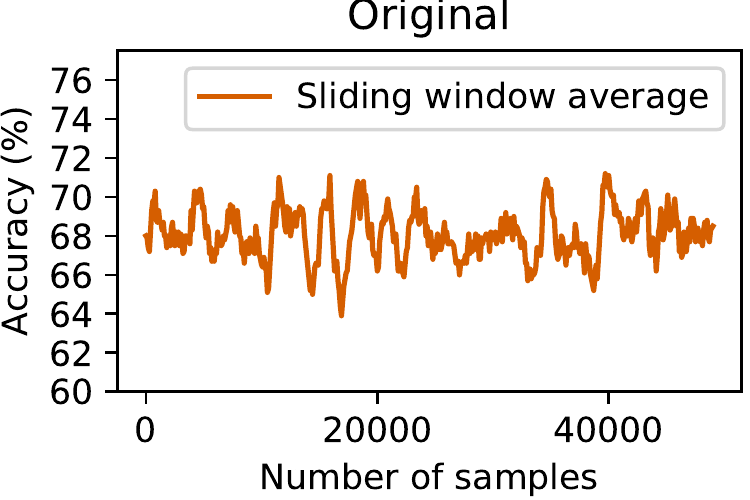}
		\includegraphics[width=0.325\textwidth]{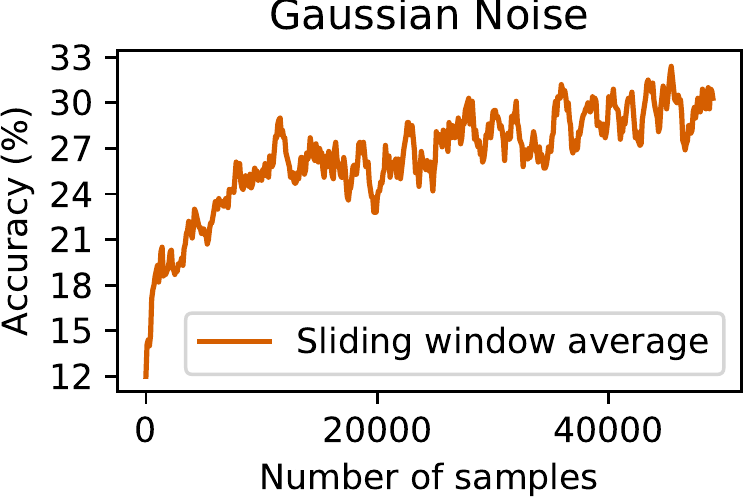}
		\includegraphics[width=0.325\textwidth]{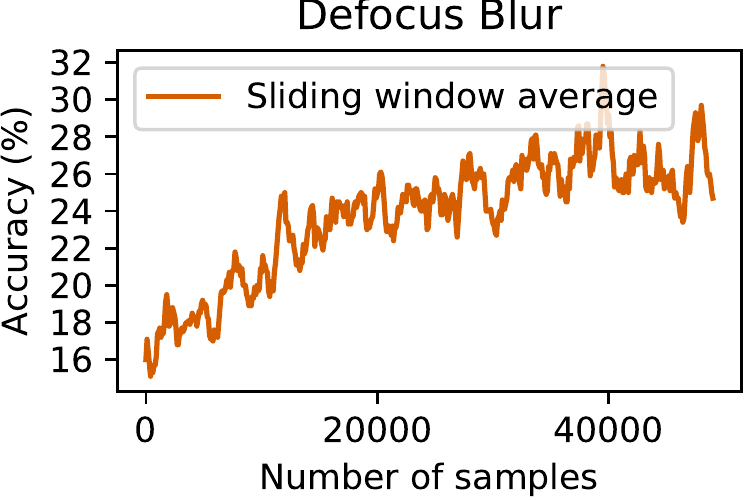}
	\end{subfigure}
	\begin{subfigure}[b]{1.0\textwidth}
			\includegraphics[width=0.325\textwidth]{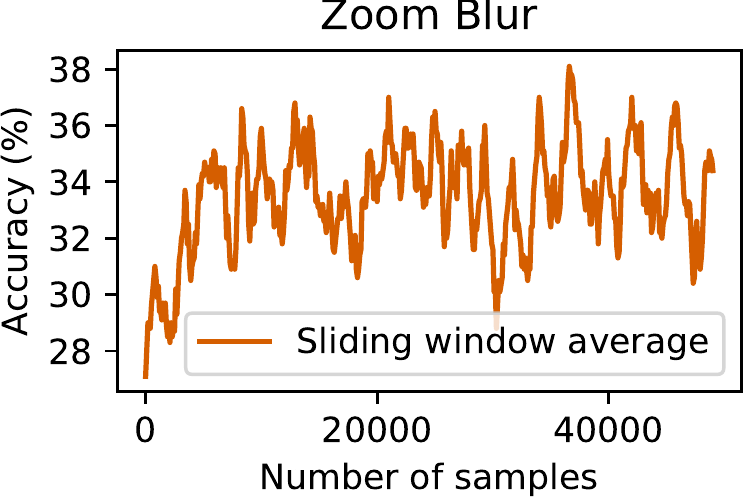}
			\includegraphics[width=0.325\textwidth]{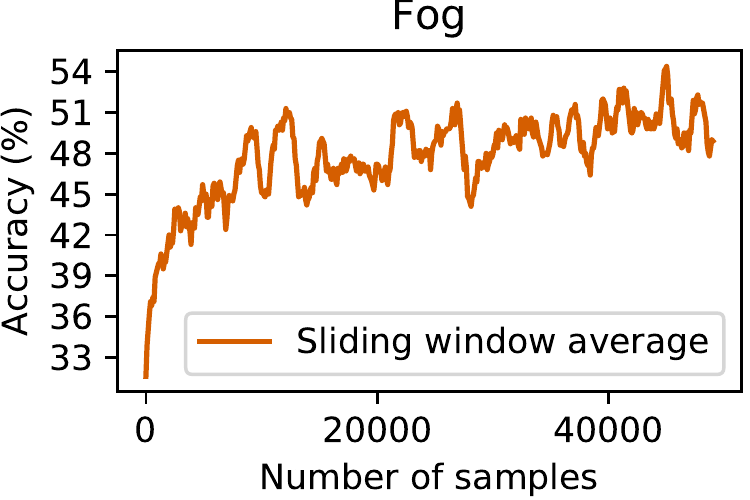}
			\includegraphics[width=0.325\textwidth]{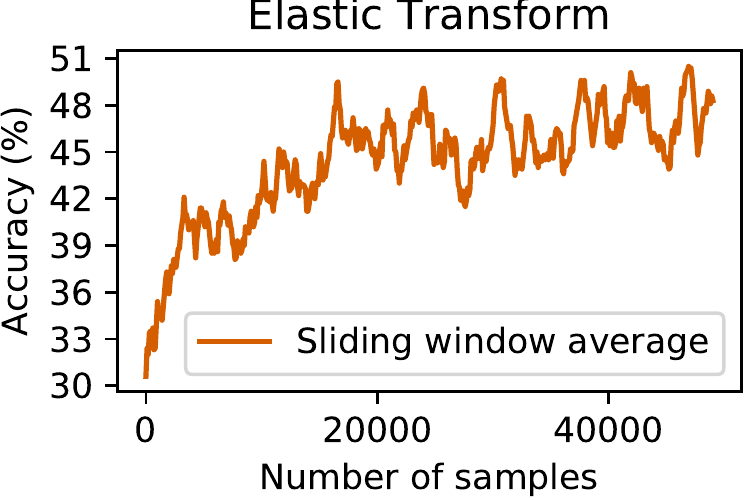}
	\end{subfigure}
	\caption{
		{\bf Test accuracy (\%) on ImageNet-C with level 5 corruptions.} Upper panel: Our approaches, TTT and TTT-Online, show significant improvements in all corruption types over the two baselines. 
		Lower panel: We show the accuracy of TTT-Online as the average over a sliding window of 100 samples; TTT-Online generalizes better as more samples are evaluated (x-axis), without hurting on the original distribution. We use accuracy instead of error here because the baseline performance is very low for most corruptions.
	}
	\label{fig:cc_imagenet}
	\vspace{-1ex}
\end{figure*}
\paragraph{Optimization details.}
For joint training (\autoref{optimize_train}),
we use stochastic gradient descent with standard hyper-parameters as \citep{huang2016deep, he2016deep}.
For Test-Time Training (\autoref{optimize_test}), we use stochastic gradient descent with the learning rate set to that of the last epoch during training, which is 0.001 in all our experiments. We set weight decay and momentum to zero during Test-Time Training, inspired by practice in \citep{he2018rethinking, liu2018rethinking}. For the standard version of Test-Time Training, we take ten gradient steps, using batches independently generated by the same image.
For online version of Test-Time Training, we take only one gradient step given each new image. We use random crop and random horizontal flip for data augmentation. See \Cref{computational} of the appendix for computational aspects of our method.
In all the tables and figures, \emph{object recognition task only} refers to the plain ResNet model (using GN, unless otherwise specified);
\emph{joint training}
refers to the model jointly trained on both the main task and the self-supervised task, fixed at test time; this has been proposed as the method in \citet{hendrycks2019using};
\emph{Test-Time Training (TTT)}  refers to the standard version described \autoref{method};
and \emph{online Test-Time Training (TTT-Online)}  refers to the online version that does not discard $\vtheta(x_t)$ for $x_t$ arriving sequentially from the same distribution.
Performance for TTT-Online is calculated as the average over the entire test set; we always shuffle the test set before TTT-Online to avoid ordering artifacts.

\subsection{Object Recognition on Corrupted Images}
\label{results_cc}
\citet{hendrycks2019benchmarking} propose to benchmark robustness of object recognition with 15 types of corruptions
from four broad categories: noise, blur, weather and digital.  Each corruption type comes in five levels of severity, with level 5 the most severe (details and sample images in the appendix).
The corruptions are simulated to mimic real-world corruptions as much as possible on copies of the test set for both CIFAR-10 and ImageNet. The new test sets are named as CIFAR-10-C and ImageNet-C, respectively. 
In the proposed benchmark, training should be done on the original training set, and the diversity of corruption types should make it difficult for any methods to work well across the board if it relies too much on corruption specific knowledge. For online Test-Time Training, we take the entire test set as a stream of incoming images, and update and test on each image in an online manner as it arrives.

\paragraph{CIFAR-10-C.}
Our results on the level 5 corruptions (most severe) are shown in \autoref{fig:cc}.
The results on levels 1-4 are shown in
\Cref{results_additional} in appendix. 
Across all five levels and 15 corruption types, both standard and online versions of Test-Time Training improve over the object recognition task only baseline by a large margin. 
The standard version always improves over joint training, and the online version often improves significantly ($>$10\%) over joint training and never hurts by more than 0.2\%. 
Specifically, TTT-Online contributes $>$24\% on the three noise types and 38\% on pixelation. 
For a learning problem with the seemingly unstable setup that abuses a single image, this kind of consistency is rather surprising.

The baseline ResNet-26 with object recognition task only has error 8.9\% on the original test set of CIFAR-10. The joint training baseline actually improves performance on the original to 8.1\%. More surprisingly, unlike many other methods that trade off original performance for robustness, Test-Time Training further improves on the original test set by 0.2\% consistently over multiple independent trials. This suggests that our method does not choose between specificity and generality.

\newpage
Separate from our method, it is interesting to note that joint training consistently improves over the single-task baseline, as discovered by \citet{hendrycks2019using}.
\citet{hendrycks2019benchmarking} have also experimented with various other training methods on this benchmark, and point to Adversarial Logit Pairing (ALP) \citep{kannan2018adversarial} as the most effective approach. 
Results of this additional baseline on all levels of CIFAR-10-C are shown in the appendix, along with its implementation details. While surprisingly robust under some of the most severe corruptions (especially the three noise types), ALP incurs a much larger error (by a factor of two) on the original distribution and some corruptions (e.g. all levels of contrast and fog), and hurts performance significantly when the corruptions are not as severe (especially on levels 1-3); 
this kind of tradeoff is to be expected for methods based on adversarial training.

\begin{table*}[ht]
	\footnotesize
	\begin{center}
		{
			\setlength\tabcolsep{4.5pt}
			\renewcommand{\arraystretch}{1.5}
			\begin{tabular}{c|c|c|c|c|c|c|c|c|c|c|c|c|c|c|c|c}
				\hline
				& orig& gauss& shot& impul& defoc& glass& motn& zoom& snow& frost& fog& brit& contr& elast& pixel& jpeg\\
				\hline
				TTT-Online & \textbf{8.2}& \textbf{25.8}& \textbf{22.6}& 30.6& \textbf{14.6}& \textbf{34.4}& \textbf{18.3}& \textbf{17.1}& \textbf{20.0}& \textbf{18.0}& \textbf{16.9}& \textbf{11.2}& 15.6& \textbf{21.6}& \textbf{18.1}& \textbf{21.2}\\
				\hline
				UDA-SS & 9.0& 28.2& 26.5& \textbf{20.8}& 15.6& 43.7& 24.5& 23.8& 25.0& 24.9& 17.2& 12.7& \textbf{11.6}& 22.1& 20.3& 22.6\\
				\hline
			\end{tabular}
			\renewcommand{\arraystretch}{1}
		}
	\end{center}
	\caption{
		{\bf Test error (\%) on CIFAR-10-C with level 5 corruption.} Comparison between online Test-Time Training (TTT-Online) and unsupervised domain adaptation by self-supervision (UDA-SS)~\citep{sun2019uda} with access to the entire (unlabeled) test set during training. We highlight the lower error in bold. We have abbreviated the names of the corruptions, in order: original test set, Gaussian noise, shot noise, impulse noise, defocus blur, glass blue, motion blur, zoom blur, snow, frost, fog, brightness, contrast, elastic transformation, pixelation, and JPEG compression. 
		The reported numbers for TTT-Online are the same as in \autoref{fig:cc}.
		See complete table in \autoref{table:full_c10c}.
	}
\label{table:uda}
\end{table*} 
\begin{figure*}[ht]
	\centering
	\includegraphics[width=0.33\textwidth]{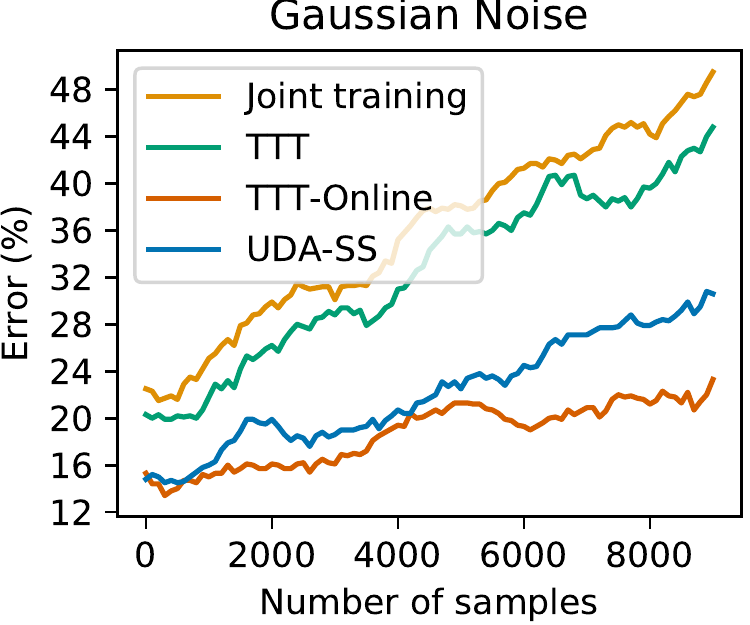}
	\includegraphics[width=0.33\textwidth]{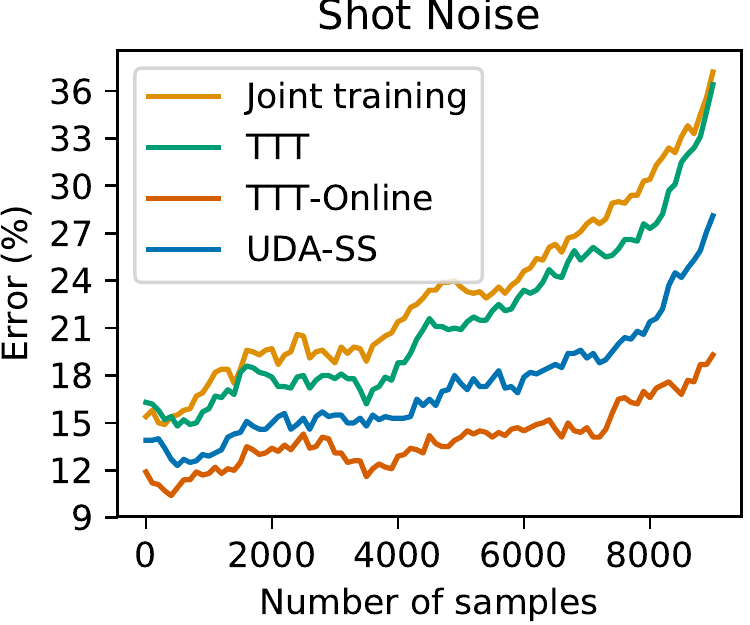}
	\includegraphics[width=0.33\textwidth]{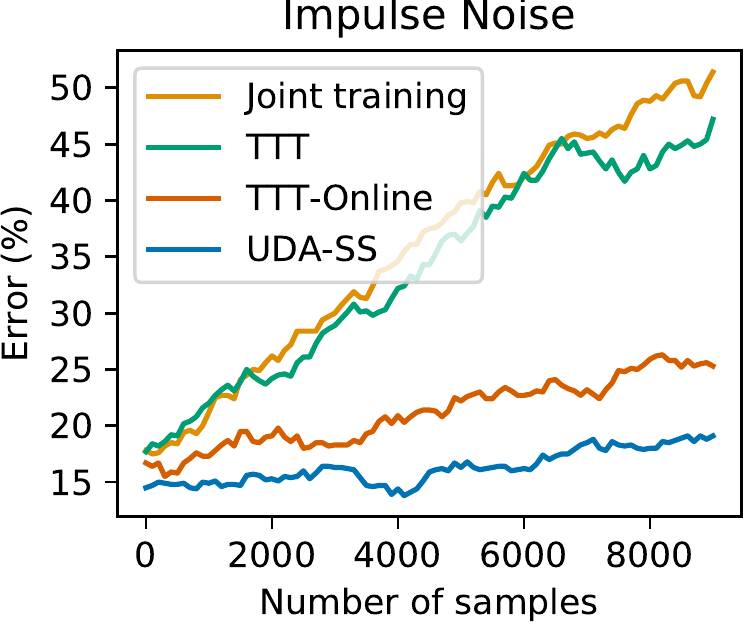}
    
	\caption{
		{\bf Test error (\%) on CIFAR-10-C, for the three noise types, with gradually changing distribution.}
		The distribution shifts are created by increasing the standard deviation of each noise type from small to large, the further we go on the x-axis. As the samples get noisier, all methods suffer greater errors the more we evaluate into the test set, but online Test-Time Training (TTT-Online) achieves gentler slopes than joint training. For the first two noise types, TTT-Online also achieves better results over unsupervised domain adaptation by self-supervision (UDA-SS)~\citep{sun2019uda}.
	}
	\label{fig:smooth}
\end{figure*}

\paragraph{ImageNet-C.}
Our results on the level 5 corruptions (most severe) are shown in \autoref{fig:cc_imagenet}. We use accuracy instead of error for this dataset 
because the baseline performance is very low for most corruptions.
The general trend is roughly the same as on CIFAR-10-C. The standard version of TTT always improves over the baseline and joint training, while the online version only hurts on the original by 0.1\% over the baseline, but significantly improves (by a factor of more than three) on many of the corruption types.

In the lower panel of \autoref{fig:cc_imagenet}, we visualize how the accuracy (averaged over a sliding window) of the online version changes as more images are tested.
Due to space constraints, we show this plot on the original test set, as well as every third corruption type, following the same order as in the original paper. 
On the original test set, there is no visible trend in performance change after updating on the 50,000 samples.
With corruptions, accuracy has already risen significantly after 10,000 samples, but is still rising towards the end of the 50,000 samples, indicating room for additional improvements if more samples were available.
Without seeing a single label, TTT-Online behaves as if we were training on the test set from the appearance of the plots. 

\paragraph{Comparison with unsupervised domain adaptation.}
\autoref{table:uda} empirically compares online Test-Time Training (TTT-Online) with unsupervised domain adaptation through self-supervision (UDA-SS)~\citep{sun2019uda}, which is similar to our method in spirit but is designed for the setting of unsupervised domain adaptation (\Cref{related} provides a survey of other related work in this setting). 
Given labeled data from the training distribution and unlabeled data from the test distribution, UDA-SS hopes to find an invariant representation that extracts useful features for both distributions by learning to perform a self-supervised task, specifically rotation prediction, simultaneously on data from both. It then learns a labeling function on top of the invariant representation using the labeled data. In our experiments, the unlabeled data given to UDA-SS is the \emph{entire test set itself} without the labels.

Because TTT-Online can only learn from the unlabeled test samples that have already been evaluated on, it is given less information than UDA-SS at all times. In this sense, UDA-SS should be regarded as an oracle rather than a baseline.
Surprisingly, TTT-Online outperforms UDA-SS on 13 out of the 15 corruptions as well as the original distribution.
Our explanation is that UDA-SS has to find an invariant representation for both distributions, while TTT-Online only adapts the representation to be good for the current test distribution. That is, TTT-Online has the flexibility to  forget the training distribution representation, which is no longer relevant. This suggests that in our setting, forgetting is not harmful and perhaps should even be taken advantage of.

\paragraph{Gradually changing distribution shifts.}
In our previous experiments, we have been evaluating the online version under the assumption that the test inputs $x_t$ for $t=1...n$ are all sampled from the same test distribution $Q$, which can be different from the training distribution $P$. 
This assumption is indeed satisfied for i.i.d. samples from a shuffled test set.
But here we show that this assumption can in fact be relaxed to allow $x_t\sim Q_t$, where $Q_t$ is close to $Q_{t+1}$ (in the sense of distributional distance).
We call this the assumption of gradually changing distribution shifts. We perform experiments by simulating such distribution shifts on the three noise types of CIFAR-10-C. For each noise type, $x_t$ is corrupted with standard deviation $\sigma_t$, and $\sigma_1,...,\sigma_n$ interpolate between the standard deviation of level 1 and level 5. So $x_t$ is more severely corrupted as we evaluate further into the test set and $t$ grows larger.
As shown in \autoref{fig:smooth}, 
TTT-Online still improves upon joint training (and our standard version) with this relaxed assumption, and even upon UDA-SS for the first two noise types.

\begin{table*}[t!]
\begin{center}
{
\setlength\tabcolsep{4pt}
\begin{tabular}{c|ccccccc|c}
\toprule
Accuracy (\%)
		&Airplane	&Bird	&Car &Dog &Cat &Horse	&Ship  	&\textbf{Average}\\
\midrule
{\small Object recognition task only} & 67.9 & 35.8 & 42.6 & 14.7 & 52.0 & 42.0 & 66.7
&\textbf{41.4}\\
\midrule
{\small Joint training {\scriptsize \citep{hendrycks2019using}}} & 70.2 & 36.7 & 42.6 & 15.5 & 52.0 & 44.0 & 66.7
&\textbf{42.4}\\
\midrule
TTT (standard version) & 70.2 & 39.2 & 42.6 & 21.6 & 54.7 & 46.0 & 77.8
&\textbf{45.2}\\
\midrule 
TTT-Online & 70.2 & 39.2 & 42.6 & 22.4 & 54.7 & 46.0 & 77.8
&\textbf{45.4}\\
\bottomrule
\end{tabular}
}
\end{center}
\caption{
Class-wise and average classification accuracy (\%) on CIFAR classes in VID-Robust, adapted from \cite{shankar2019systematic}. Test-Time Training (TTT) and online Test-Time Training (TTT-Online) improve over the two baselines on average, and by a large margin on ``ship'' and ``dog'' classes where the rotation task is more meaningful than in classes like ``airplane'' (sample images in \autoref{vid_samples}). 
}
\label{table.cifar7}
\vspace{-1ex}
\end{table*}

\subsection{Object Recognition on Video Frames}
\label{ivc}
The Robust ImageNet Video Classification (VID-Robust) dataset was developed by \citet{shankar2019systematic} from the ImageNet Video detection dataset \citep{ILSVRC15}, to demonstrate how deep models for object recognition trained on ImageNet (still images) fail to adapt well to video frames. The VID-Robust dataset contains 1109 sets of video frames in 30 classes; each set is a short video clip of frames that are similar to an anchor frame. Our results are reported on the anchor frames. To map the 1000 ImageNet classes to the 30 VID-Robust classes, we use the max-conversion function in \citet{shankar2019systematic}.
Without any modifications for videos, we apply our method to VID-Robust on top of the same ImageNet model as in the previous subsection. 
Our classification accuracy is reported in \autoref{vidtable}. 

In addition, we take the seven classes in VID-Robust that overlap with CIFAR-10, and re-scale those video frames to the size of CIFAR-10 images, as a new test set for the model trained on CIFAR-10 in the previous subsection. Again, we apply our method to this dataset without any modifications.
Our results are shown in \autoref{table.cifar7}, with a breakdown for each class. 
Noticing that Test-Time Training does not improve on the airplane class, we inspect some airplane samples (\autoref{vid_samples}), and observe black margins on two sides of most images, which provide a trivial hint for rotation prediction. 
In addition, given an image of airplanes in the sky, it is often impossible even for humans to tell if it is rotated.
This shows that our method requires the self-supervised task to be both well defined and non-trivial.

\subsection{CIFAR-10.1: Unknown Distribution Shifts}
\label{cifar10_new}
CIFAR-10.1 \citep{recht2018cifar} is a new test set of size 2000 modeled after CIFAR-10, with the exact same classes and image dimensionality, following the dataset creation process documented by the original CIFAR-10 paper as closely as possible.
The purpose is to investigate the distribution shifts present between the two test sets, and the effect on object recognition.
All models tested by the authors suffer a large performance drop on CIFAR-10.1 comparing to CIFAR-10,
even though there is no human noticeable difference, and both have the same human accuracy.
This demonstrates how insidious and ubiquitous distribution shifts are, even when researchers strive to minimize them.

\begin{table}
	\vspace{-1ex}
	\begin{center}{
			\begin{tabular}{c|c}  \toprule
				Method 									& Accuracy (\%) \\	\midrule
				{\small Object recognition task only} 								& 62.7\\  \midrule
				{\small Joint training {\scriptsize \citep{hendrycks2019using}}}						& 63.5\\  \midrule
				TTT	(standard version)	& 63.8\\  \midrule
				TTT-Online		& 64.3\\  \bottomrule
			\end{tabular}
			\caption{Test accuracy (\%) on VID-Robust dataset~\cite{shankar2019systematic}. 
			TTT and TTT-Online improve over the baselines.
			}\label{vidtable}
	}\end{center}
\end{table} 

\begin{table}
\begin{center}{
\begin{tabular}{c|c}  \toprule
Method 									& Error (\%) \\	\midrule
{\small Object recognition task only} 								
& 17.4\\  \midrule
{\small Joint training {\scriptsize \citep{hendrycks2019using}}}						
& 16.7\\  \midrule
TTT	(standard version)		& 15.9\\  \bottomrule
\end{tabular}
\caption{Test error (\%) on CIFAR-10.1~\cite{recht2018cifar}. TTT is the first method to improve the performance of an existing model on this new test set.}
\label{cifar10_new_table}
}\end{center}
\vspace{-3ex}
\end{table}

The distribution shifts from CIFAR-10 to CIFAR-10.1 pose an extremely difficult problem, and no prior work has been able to improve the performance of an existing model on this new test set, 
probably because:
1) researchers cannot even identify the distribution shifts, let alone describe them mathematically; 
2) the samples in CIFAR-10.1 are only revealed at test time; and even if they were revealed during training, the distribution shifts are too subtle, and the sample size is too small, for domain adaptation~\citep{recht2018cifar}.

On the original CIFAR-10 test set, the baseline with only object recognition has error 8.9\%,
and with joint training has 8.1\%;
comparing to the first two rows of \autoref{cifar10_new_table}, 
both suffer the typical performance drop (by a factor of two). 
TTT yields an improvement of 0.8\%
(relative improvement of 4.8\%) over joint training.
We recognize that this improvement is small relative to the performance drop, but see it as an encouraging first step for this very difficult problem.

\begin{figure*}
	\centering
	\includegraphics[width=0.48\textwidth]{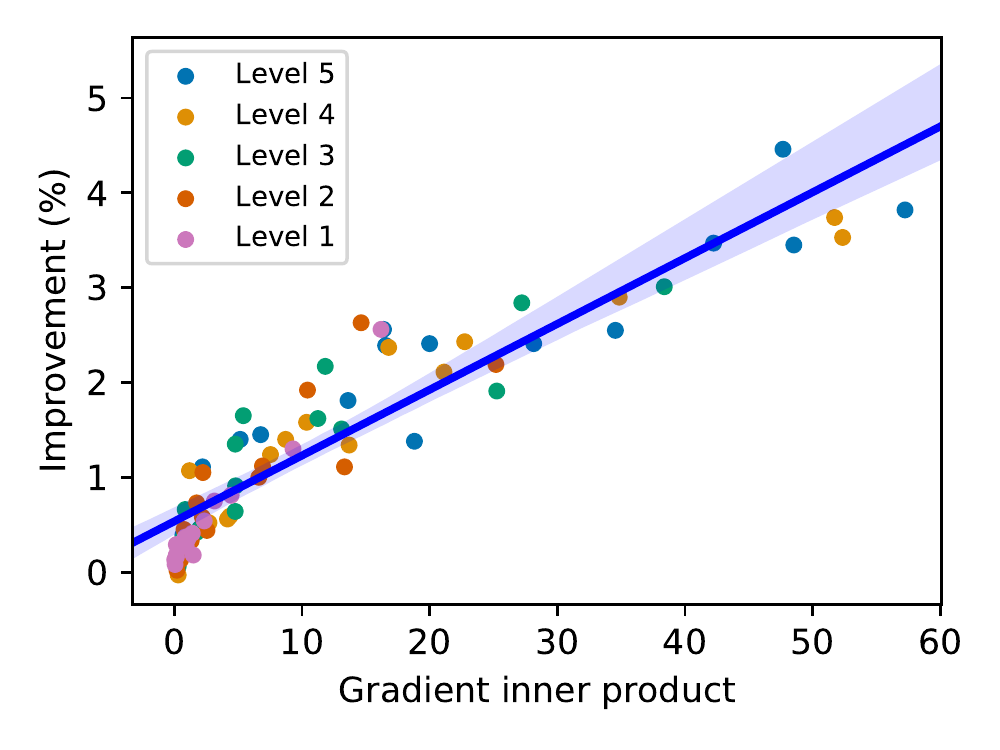}
	\includegraphics[width=0.48\textwidth]{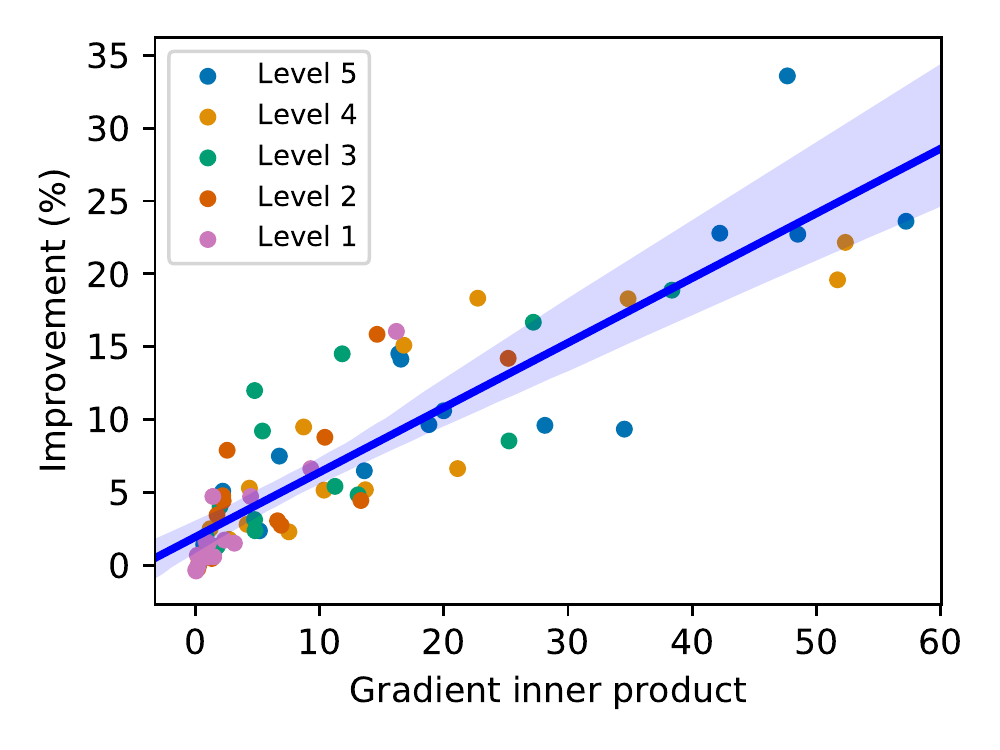}
	\caption{
		Scatter plot of the inner product between the gradients (on the shared feature extractor $\vtheta_e$) of
		the main task $l_m$ and the self-supervised task $l_e$, and the improvement in test error (\%) from Test-Time Training, 
		for the standard (left) and online (right) version. 
		Each point is the average over a test set, and each scatter plot has 75 test sets, from all 15 types of corruptions over five levels as described in \autoref{results_cc}.
		The blue lines and bands are the best linear fits and the 99\% confidence intervals.
		The linear correlation coefficients are $0.93$ and $0.89$ respectively, indicating strong positive correlation between the two quantities, as suggested by \Cref{main_theorem}.
	}
	\label{gradient_improve}
\end{figure*}

\section{Theoretical Results}
\label{theory}
This section contains our preliminary study of when and why Test-Time Training is expected to work.
For convex models, we prove that positive gradient correlation between the loss functions leads to better performance on the main task after Test-Time Training.
Equipped with this insight, we then empirically
demonstrate that gradient correlation governs the success of Test-Time Training on the deep learning model discussed in \Cref{results}.

Before stating our main theoretical result, we first illustrate the general intuition with a toy model.
Consider a regression problem where $x\in\R^d$ denotes the input,  $y_1\in\R$ denotes the label, and the objective is the square loss
$(\hat{y}-y_1)^2/2$ for a prediction $\hat{y}$.  
Consider a two layer linear network parametrized by $\mA\in\R^{h\times d}$ and $\vv \in\R^h$ (where $h$
stands for the hidden dimension).  
The prediction according to this model
is $\hat{y}=\vv^\top \mA x$, and the main task loss is
\begin{equation}
    l_m(x, y_1; \mA, \vv) = \frac{1}{2}\left(y_1 - \vv^ \top \mA x\right)^2.
\end{equation}
In addition, consider a self-supervised regression task that also uses the square loss and automatically generates a label $y_s$ for $x$. 
Let the self-supervised head be parametrized by $\vw\in\R^h$.  Then the self-supervised task loss is
\begin{equation}
    l_s(x, y_2; \mA, \vw) = \frac{1}{2}\left(y_2 - \vw^\top\mA x\right)^2.
\end{equation}
Now we apply Test-Time Training to update the shared feature extractor $\mA$ by
one step of gradient descent on $l_s$, which we can compute with $y_2$ known.
This gives us
\begin{equation}
    \mA' \leftarrow \mA - \eta\left( y_2-\vw^\top\mA x \right) \left(-\vw x^\top\right),
\end{equation}
where $\mA'$ is the updated matrix and $\eta$ is the learning rate.
If we set $\eta = \eta^*$ where
\begin{equation}
\label{magic_lr}
\eta^* = \frac{y_1 - \vv^\top \mA x}{\left( y_2 - \vw^\top \mA x \right)
	\vv^\top \vw x^\top x},
\end{equation}
then with some simple algebra, it is easy to see that the main task loss 
$l_m(x, y_1; \mA', \vv) = 0$.  
Concretely, Test-Time Training drives the main task loss down to zero with a single gradient step for a carefully chosen learning rate. 
In practice, this learning rate is unknown since it depends on the unknown $y_1$.
However, since our model is convex, as long as $\eta\opt$ is positive, it suffices to set $\eta$ to be a small positive constant (see details in the appendix).
If $x \neq 0$, one sufficient condition for $\eta^*$ to be positive (when neither loss is zero) is to have
\begin{align}
\label{equal_sign}
    \sign\left( y_1 - \vv^\top \mA x \right) = \sign\left( y_2 - \vw^\top \mA x \right)\\
    \text{and}\quad
    \vv^\top \vw > 0 \;.
\end{align}
For our toy model, both parts of the condition above have an intuition interpretation.
The first part says that the mistakes should be correlated, in the sense
that predictions from both tasks are mistaken in the same direction. The second
part, $v^\top \vw > 0$, says that the decision boundaries on the feature
space should be correlated. In fact, these two parts hold iff.
$\langle \grad l_m(\mA), \grad l_s(\mA)\rangle > 0$ 
(see a simple proof of this fact in the appendix). To summarize, if the gradients have
positive correlation, Test-Time Training is guaranteed to reduce the main task loss.
Our main theoretical result extends this to general smooth and convex loss functions.

\newpage
{\theorem	\label{main_theorem}
Let $l_m(x, y; \vtheta)$ denote the main task loss on test instance $x, y$ with
parameters $\vtheta$, and $l_s(x; \vtheta)$ the self-supervised task loss that only depends on $x$.
Assume that for all $x, y$,
$l_m(x, y; \vtheta)$ is differentiable, convex and $\beta$-smooth
in $\vtheta$,
and both
$\norm{\grad l_m(x, y; \vtheta)}, \norm{\grad l_s(x, \vtheta)} \leq G$ 
for all $\vtheta$.
With a fixed learning rate $\eta = \frac{\eps}{\beta G^2}$, 
for every $x, y$ such that 
\begin{align}
\langle \grad l_m(x, y; \vtheta), \grad l_s(x; \vtheta) \rangle  > \eps,
\end{align}
we have
\begin{align}
l_m(x, y; \vtheta)
>  l_m(x, y; \vtheta(x)),
\end{align}
where $\vtheta(x) = \vtheta - \eta \grad l_s(x; \vtheta)$ i.e. Test-Time Training with one step of gradient descent.
}

The proof uses standard techniques in optimization, and is left for
the appendix.  Theorem 1 reveals gradient correlation as a determining factor of the success of Test-Time Training in the smooth and convex case.
In \autoref{gradient_improve}, we empirically show that our insight also holds for non-convex loss functions,
on the deep learning model and across the diverse set of corruptions considered in \Cref{results}; stronger gradient correlation clearly indicates more performance improvement over the baseline.

\section{Related Work}
\label{related}

\paragraph{Learning on test instances.}
\citet{shocher2018zero} provide a key inspiration for our work by showing that image super-resolution could be learned at test time simply by trying to upsample a downsampled version of the input image. More recently, \citet{bau2019semantic} improve photo manipulation by adapting a pre-trained GAN to the statistics of the input image.   One of the earlier examples of this idea comes from \citet{jain2011online}, who improve Viola-Jones face detection \citep{viola2001rapid} by bootstrapping the more difficult faces in an image from the more easily detected faces in that same image. 
The online version of our algorithm is inspired by the work of \citet{mullapudi2018online}, which makes video segmentation more efficient by using a student model that learns online from a teacher model.
The idea of online updates has also been used in \citet{kalal2011tracking} for tracking and detection.
A recent work in echocardiography \citep{zhu2019neural} improves the deep learning model that tracks myocardial motion and cardiac blood flow with sequential updates.
Lastly, we share the philosophy of transductive learning \citep{vapnik2013nature, gammerman1998learning}, but have little in common with their classical algorithms;
recent work by \citet{nilesh} theoretically explores this for linear prediction, in the context of debiasing the LASSO estimator.

\paragraph{Self-supervised learning} studies how to create labels from the data, by designing various pretext tasks that can learn semantic information without human annotations, 
such as context prediction \citep{doersch2015unsupervised}, solving jigsaw puzzles \citep{noroozi2016unsupervised}, colorization \citep{larsson2017colorproxy,zhang2016colorful}, noise prediction \citep{bojanowski2017unsupervised}, feature clustering \citep{caron2018deep}. 
Our paper uses rotation prediction \citep{gidaris2018unsupervised}.
\citet{asano2019surprising} show that self-supervised learning on only a single image, surprisingly, can produce low-level features that generalize well.
Closely related to our work, \citet{hendrycks2019using} propose that jointly training a main task and a self-supervised task (our joint training baseline in \Cref{results}) can improve robustness on the main task.
The same idea is used in few-shot learning  \citep{su2019boosting}, domain generalization \citep{carlucci2019domain}, and unsupervised domain adaptation \citep{sun2019uda}.

\paragraph{Adversarial robustness}
studies the robust risk
$
R_{P, \Delta}(\vtheta) = \E_{x,y\sim P} \max_{\delta \in \Delta}  l(x+\delta, y; ~\vtheta),
$
where $l$ is some loss function, and $\Delta$ is the set of perturbations; $\Delta$ is often chosen as the $L_p$ ball, for $p\in \{1, 2, \infty\}$.
Many popular algorithms formulate and solve this as a robust optimization problem \citep{goodfellow2014explaining, madry2017towards, sinha2017certifying, raghunathan2018certified, wong2017provable, croce2018provable}, and the most well known technique is adversarial training.
Another line of work is based on randomized smoothing \citep{cohen2019certified, salman2019provably}, while some other approaches, such as input transformations
\citep{guo2017countering, song2017pixeldefend}, are shown to be less effective \citep{athalye2018obfuscated}.
There are two main problems with the approaches above.
First, all of them can be seen as \emph{smoothing} the decision boundary.
This establishes a theoretical tradeoff between accuracy and robustness \citep{tsipras2018robustness, zhang2019theoretically}, which we also observe empirically with our adversarial training baseline in \Cref{results}.
Intuitively, the more diverse $\Delta$ is, the less effective this \emph{one-boundary-fits-all} approach can be for a particular element of $\Delta$.
Second, adversarial methods rely heavily on the mathematical structure of $\Delta$, which might not accurately model perturbations in the real world. 
Therefore, generalization remains hard outside of the $\Delta$ we know in advance or can mathematically model, especially for non-adversarial distribution shifts.
Empirically, \citet{kang2019transfer} shows that robustness for one $\Delta$ might not transfer to another, and training on the $L_\infty$ ball actually hurts robustness on the $L_1$ ball.

\paragraph{Non-adversarial robustness} studies the effect of corruptions, perturbations, out-of-distribution examples, and real-world distribution shifts
\citep{hendrycks2019improving, hendrycks2019using, hendrycks2018using, hendrycks2016baseline}.
\citet{geirhos2018generalisation} show that training on images corrupted by Gaussian noise makes deep learning models robust to this particular noise type, but does not improve performance on images corrupted by another noise type e.g. salt-and-pepper noise.

\paragraph{Unsupervised domain adaptation}
(a.k.a. transfer learning) studies the problem of distribution shifts, when an unlabeled dataset from the test distribution (target domain) is available at training time, 
in addition to a labeled dataset from the training distribution (source domain)
\citep{chen2011co, gong2012geodesic, long2015learning, ganin2016domain, long2016unsupervised, tzeng2017adversarial,hoffman2017cycada,csurka2017domain, chen2018adversarial}.
The limitation of the problem setting, however, is that generalization might only be improved for this specific test distribution, which can be difficult to anticipate in advance.
Prior work try to anticipate broader distributions by using multiple and evolving domains \citep{hoffman2018algorithms, hoffman2012discovering, hoffman2014continuous}.
Test-Time Training does not anticipate any test distribution, by changing the setting of unsupervised domain adaptation, while taking inspiration from its algorithms.
Our paper is a follow-up to \citet{sun2019uda}, which we explain and empirically compare with in \Cref{results}.
Our update rule can be viewed as performing \emph{one-sample unsupervised domain adaptation} on the fly, with the caveat that standard domain adaptation techniques might become ill-defined when there is only one sample from the target domain.

\paragraph{Domain generalization} studies the setting where a meta distribution generates multiple environment distributions, some of which are available during training (source), while others are used for testing (target)
\citep{li2018deep, shankar2018generalizing, muandet2013domain, balaji2018metareg, ghifary2015domain, motiian2017unified, li2017deeper, gan2016learning}. 
With only a few environments, information on the meta distribution is often too scarce to be helpful,
and with many environments, we are back to the i.i.d. setting where each environment can be seen as a sample, and a strong baseline is to simply train on all the environments \citep{li2019episodic}.
The setting of domain generalization is limited by the inherent tradeoff between specificity and generality of a fixed decision boundary, and the fact that generalization is again elusive outside of the meta distribution i.e. the actual $P$ learned by the algorithm.

\paragraph{One (few)-shot learning} studies how to learn a new task or a new classification category using only one (or a few) sample(s), on top of a general representation that has been learned on diverse samples 
\citep{snell2017prototypical, vinyals2016matching, fei2006one, ravi2016optimization, li2017meta, finn2017model, gidaris2018dynamic}.
Our update rule can be viewed as performing \emph{one-shot self-supervised learning} and can potentially be improved by progress in one-shot learning.

\paragraph{Continual learning} (a.k.a. learning without forgetting) studies the setting where a model is made to learn a sequence of tasks, and not forget about the earlier ones while training for the later \citep{li2017learning, lopez2017gradient, kirkpatrick2017overcoming, santoro2016meta}. In contrast, with Test-Time Training, we are not concerned about forgetting the past test samples since they have already been evaluated on; and if a past sample comes up by any chance, it would go through Test-Time Training again. In addition, the impact of forgetting the training set is minimal, because both tasks have already been jointly trained.

\paragraph{Online learning} (a.k.a. online optimization) is a well-studied area of learning theory
\citep{shalev2012online, hazan2016introduction}. 
The basic setting repeats the following: receive $x_t$, predict $\hat{y}_t$, receive $y_t$ from a worst-case oracle, and learn.
Final performance is evaluated using the regret, which colloquially translates to how much worse the online learning algorithm performs in comparison to the best fixed model in hindsight.
In contrast, our setting never reveals any $y_t$ during testing even for the online version, so we do not need to invoke the concept of the worst-case oracle or the regret.
Also, due to the lack of feedback from the environment after predicting, our algorithm is motivated to learn (with self-supervision) before predicting $\hat{y}_t$ instead of after.
Note that some of the previously covered papers \cite{hoffman2014continuous, jain2011online, mullapudi2018online} use the term ``online learning'' outside of the learning theory setting, so the term can be overloaded.

\section{Discussion}
The idea of test-time training also makes sense for other tasks, such as segmentation and detection, and in other fields, such as speech recognition and natural language processing. 
For machine learning practitioners with prior domain knowledge in their respective fields, their expertise can be leveraged to design better special-purpose self-supervised tasks for test-time training.
Researchers for general-purpose self-supervised tasks can also use  test-time training as an evaluation benchmark, in addition to the currently prevalent benchmark of pre-training and fine-tuning.

More generally, we hope this paper can encourage researchers to abandon the self-imposed constraint of a fixed decision boundary for testing, or even the artificial division between training and testing altogether. 
Our work is but a small step toward a new paradigm where much of the learning happens {\em after} a model is deployed.

\paragraph{Acknowledgements.}
This work is supported by NSF grant 1764033, DARPA and Berkeley DeepDrive.
This paper took a long time to develop, and benefited from conversations with many of our colleagues, including
Ben Recht and his students Ludwig Schmidt, Vaishaal Shanker and Becca Roelofs;
Ravi Teja Mullapudi, Achal Dave and Deva Ramanan;
and Armin Askari, Allan Jabri, Ashish Kumar,
Angjoo Kanazawa and Jitendra Malik.

\bibliography{ttt}
\bibliographystyle{icml2020}


%
%

\renewcommand{\thesection}{A\arabic{section}}
\renewcommand{\thesubsection}{\thesection.\arabic{subsection}}

\newcommand{\beginsupplementary}{%
	\setcounter{table}{0}
	\renewcommand{\thetable}{A\arabic{table}}%
	\setcounter{figure}{0}
	\renewcommand{\thefigure}{A\arabic{figure}}%
	\setcounter{section}{0}
}
\beginsupplementary
\newcommand{\suptitl}{Appendix: Test-Time Training with Self-Supervision\\
	for Generalization under Distribution Shifts}
\newcommand{\suptitlrunning}{Test-Time Training with Self-Supervision
	for Generalization under Distribution Shifts}
\icmltitlerunning{\suptitlrunning}
\twocolumn[
\icmltitle{\suptitl}
\vskip 0.3in
]

\section{Informal Discussion on Our Variable Decision Boundary}
\label{vdb}
In the introduction, we claim that in traditional supervised learning $\vtheta$ gives a fixed decision boundary, while our $\vtheta$ gives a variable decision boundary. Here we informally discuss this claim.

Denote the input space $\scrX$ and output space $\scrY$. 
A decision boundary is simply a mapping $f: \scrX \rightarrow \scrY$.
Let $\Theta$ be a model class e.g $\mathbb{R}^d$.
Now consider a family of parametrized functions $g_\vtheta: \scrX \rightarrow \scrY$, where $\vtheta\in \Theta$.
In the context of deep learning, $g$ is the neural network architecture and $\vtheta$ contains the parameters. 
We say that $f$ is a fixed decision boundary w.r.t. $g$ and $\Theta$
if there exists $\vtheta\in\Theta$ s.t. $f(x) = g_\vtheta(x)$ for every $x\in\scrX$,
and a variable decision boundary if for every $x\in\scrX$, there exists $\vtheta\in\Theta$ s.t. $f(x) = g_\vtheta(x)$.
Note how selection of $\vtheta$ can depend on $x$ for a variable decision boundary, and cannot for a fixed one. It is then trivial to verify that our claim is true under those definitions.

A critical reader might say that with an arbitrarily large model class, can't every decision boundary be fixed?
Yes, but this is not the end of the story.
Let $d = \dim(\scrX) \times \dim(\scrY)$, and consider the enormous model class $\Theta' = \R^d$ which is capable of representing all possible mappings between $\scrX$ and $\scrY$.
Let $g'_{\vtheta'}$ simply be the mapping represented by $\vtheta'\in\Theta'$.
A variable decision boundary w.r.t. $g$ and $\Theta$ then indeed must be a fixed decision boundary w.r.t. $g'$ and $\Theta'$, but we would like to note two things.
First, without any prior knowledge, generalization in $\Theta'$ is impossible with any finite amount of training data; 
reasoning about  $g'$ and $\Theta'$ is most likely not productive from an algorithmic point of view, and the concept of a variable decision boundary is to avoid such reasoning.
Second, selecting $\vtheta$ based on $x$ for a variable decision boundary can be thought of as ``training" on all points $x\in\R^d$; however, ``training" only happens when necessary, for the $x$ that it actually encounters. 

Altogether, the concept of a variable decision boundary is different from what can be described
by traditional learning theory. 
A formal discussion is beyond the scope of this paper and might be of interest to future work.

\section{Computational Aspects of Our Method}
\label{computational}
At test time, our method is $2~\times$
\texttt{batch\_size}
$\times$
\texttt{number\_of\_iterations} 
times slower than regular testing, 
which only performs a single forward pass for each sample. 
As the first work on Test-Time Training, this paper is not as concerned about computational efficiency as improving robustness, but here we provide two potential solutions that might be useful, but have not been thoroughly verified.
The first is to use the thresholding trick on $l_s$, introduced as a solution for the small batches problem in the method section. For the models considered in our experiments, roughly $80\%$ of the test instances fall below the threshold, so Test-Time Training can only be performed on the other $20\%$ without much effect on performance, because those $20\%$ contain most of the samples with wrong predictions.
The second is to reduce the \texttt{number\_of\_iterations} of test-time updates. For the online version, the \texttt{number\_of\_iterations} is already 1, so there is nothing to do. For the standard version, we have done some preliminary experiments setting \texttt{number\_of\_iterations} to 1 (instead of 10) and learning rate to 0.01 (instead of 0.001), and observing results almost as good as the standard hyper-parameter setting.
A more in depth discussion on efficiency is left for future works, which might, during training, explicitly make the model amenable to fast updates.

\section{Proofs}
Here we prove the theoretical results in the main paper.
\subsection{The Toy Problem}
The following setting applies to the two lemmas; this is simply the setting of our toy problem, reproduced here for ease of reference.

\pagebreak
Consider a two layer linear network parametrized by $\mA\in\R^{h\times d}$ (shared) and $\vv, \vw\in\R^h$ (fixed) for the two heads, respectively.
Denote $x\in\R^d$ the input and $y_1, y_2\in\R$ the labels for the two tasks, respectively.
For the main task loss
\begin{equation}
l_m(\mA; \vv) = \frac{1}{2}\left(y_1 - \vv^\top \mA x\right)^2,
\end{equation}
and the self-supervised task loss
\begin{equation}
l_s(\mA; \vw) = \frac{1}{2}\left(y_2 - \vw^\top \mA x\right)^2,
\end{equation}
Test-Time Training yields an updated matrix
\begin{equation}
\mA' \leftarrow \mA - \eta\left( y_2-\vw^\top \mA x \right) \left(-\vw x^\top \right),
\end{equation}
where $\eta$ is the learning rate.

{\lemma \label{lemma_lr}
Following the exposition of the main paper, let
\begin{equation}
\eta\opt  = \frac{(y_1 - v^\top A x)}{(y_2 - w^\top A x) v^\top w x^\top x}.
\end{equation}
Assume $\eta\opt \in [\epsilon, \infty)$ for some $\epsilon > 0$.
Then for any $\eta \in (0, \epsilon]$,
we are guaranteed an improvement on the main loss i.e. $l_m(\mA') < l_m(\mA)$.
}
{\proof
From the exposition of the main paper, we know that
$$l_m(\mA - \eta\opt \grad l_s\mA)) = 0,$$
which can also be derived from simple algebra.
Then by convexity, we have
\begin{align}
l_m&\paren{\mA - \eta \grad l_s(\mA)}\\
&= l_m\paren{\paren{1 - \frac{\eta}{\eta\opt}} \mA 
	+ \frac{\eta}{\eta\opt}(\mA - \eta\opt \grad l_s(\mA))}  \\
&\leq \paren{1 - \frac{\eta}{\eta\opt}} l_m(\mA) + 0 \\
&\leq \paren{1 - \frac{\eta}{\epsilon}} l_m(\mA)\\
&< l_m(\mA),
\end{align}
where the last inequality uses the assumption that $l_m(\mA)>0$, which holds
because $\eta\opt > 0$.}
{\lemma \label{lemma_sign}
Define
$\langle \mU, \mV \rangle
= \text{vec} \left(\mU\right)^\top \text{vec} \left(\mV \right)$ 
i.e. the Frobenious inner product, then
\begin{equation}
\sign \left(\eta\opt\right) = \sign \left(\langle \grad l_m(\mA), \grad l_s(\mA)\rangle\right).
\end{equation}
}
{\proof By simple algebra,
\begin{align*}
&\langle\grad  l_m(\mA), \grad l_s(\mA)\rangle \\
&= 
    \langle\left( y_1-\vv^\top \mA x \right)\left(-\vv x^\top \right),
    \left( y_2-\vw^\top \mA x \right)\left(-\vw x^\top \right)\rangle\\
&= \left( y_1-\vv^\top \mA x \right)\left( y_2-\vw^\top \mA x \right)
\Tr \left( x \vv^\top \vw x^\top \right)\\
&= \left( y_1-\vv^\top\mA x \right)\left( y_2-\vw^\top\mA x \right) 
\vv^\top \vw x^\top x,
\end{align*}
}
which has the same sign as $\eta\opt$.

\subsection{Proof of Theorem 1}
\label{main_theorem_proof}
For any $\eta$, by smoothness and convexity,
\begin{align*}
\label{proof_first_eq}
&l_m(x,y; \vtheta(x))
= l_m(x,y; \vtheta - \eta \grad l_s(x; \vtheta))\\
&\leq l_m(x,y; \vtheta) 
+ \eta \langle \grad l_m(x, y; \vtheta), \grad l_s(x, \vtheta) \rangle \\
&\qquad+ \frac{\eta^2\beta}{2} \norm{\grad l_s(x; \vtheta)}^2.
\end{align*}
Denote
$$\eta\opt = \frac{\langle \grad l_m(x, y; \vtheta), \grad l_s(x, \vtheta) \rangle}
{\beta \norm{\grad l_s(x; \vtheta)}^2}.$$ 
Then \autoref{proof_first_eq} becomes
\begin{align}
&l_m(x,y; \vtheta - \eta\opt \grad l_s(x; \vtheta))\\
&\leq l_m(x,y; \vtheta) 
- \frac{\langle \grad l_m(x, y; \vtheta), \grad l_s(x, \vtheta) \rangle^2}
{2\beta \norm{\grad l_s(x; \vtheta)}^2}.
\end{align}
And by our assumptions on the gradient norm 
and gradient inner product,
\begin{align}
\label{proof_second_eq}
l_m(x,y; \vtheta)  - l_m(x,y; \vtheta - \eta\opt \grad l_s(x; \vtheta)) \geq\frac{\eps^2}{2\beta G^2}.
\end{align}
Because we cannot observe $\eta\opt$ in practice, we instead use a fixed learning rate $\eta = \frac{\eps}{\beta G^2}$, as stated in Theorem 1.
Now we argue that this fixed learning rate still improves performance on the main task.

By our assumptions, $\eta\opt \geq \frac{\eps}{\beta G^2}$, 
so $\eta \in (0, \eta\opt]$.
Denote $\vg = \grad l_s(x; \vtheta)$, then by convexity of $l_m$,
\begin{align}
&l_m(x,y; \vtheta(x))
= l_m(x,y; \vtheta - \eta \vg)\\
&= l_m\paren{x,y; 
	\paren{1 - \frac{\eta}{\eta\opt}} \vtheta +
	\frac{\eta}{\eta\opt} \paren{\vtheta - \eta\opt \vg}} \\
&\leq \paren{1 - \frac{\eta}{\eta\opt}} l_m(x,y; \vtheta) 
+ \frac{\eta}{\eta\opt} l_m(x,y; \vtheta - \eta\opt g)
\end{align}
Combining with \autoref{proof_second_eq}, we have
\begin{align*}
l_m(x,y; \vtheta(x))
&\leq \paren{1 - \frac{\eta}{\eta\opt}} l_m(x,y; \vtheta) \\
&\qquad
+ \frac{\eta}{\eta\opt} \paren{l_m(x,y; \vtheta) - \frac{\eps^2}{2\beta G^2}} \\
&= l_m(x,y; \vtheta) - \frac{\eta}{\eta\opt}\frac{\eps^2}{2\beta G^2}
\end{align*}
Since $\eta / \eta\opt > 0$, we have shown that
\begin{align}
l_m(x,y; \vtheta)  - l_m(x,y; \vtheta(x)) > 0.
\end{align}

\pagebreak

\section{Additional Results on the Common Corruptions Dataset}
\label{results_additional}
For table aethetics, we use the following abbreviations:
B for baseline, JT for joint training, TTT for Test-Time Training standard version, and TTT-Online for online Test-Time Training i.e. the online version.

We have abbreviated the names of the corruptions, in order: original test set, Gaussian noise, shot noise, impulse noise, defocus blur, glass blue, motion blur, zoom blur, snow, frost, fog, brightness, contrast, elastic transformation, pixelation, and JPEG compression. 

\subsection{Results Using Batch Normalization}
As discussed in the results section, Batch Normalization (BN) is ineffective for small batches, which are the inputs for Test-Time Training (both standard and online version) since there is only one sample available when forming each batch; therefore, our main results are based on a ResNet using Group Normalization (GN). 
\autoref{figure:bn} and \autoref{table:bn} show results of our method on CIFAR-10-C level 5, with a ResNet using Batch Normalization (BN). These results are only meant to be a point of reference for the curious readers.

In the early stage of this project, we have experimented with two potential solutions to the small batches problem with BN.  The naive solution is to fix the BN layers during Test-Time Training. but this diminishes the performance gains since there are fewer shared parameters. 
The better solution, adopted for the results below, is hard example mining: instead of updating on all inputs, we only update on inputs that incur large self-supervised task loss $l_s$,
where the large improvements might counter the negative effects of inaccurate statistics.

Test-Time Training (standard version) is still very effective with BN. 
In fact, some of the improvements are quite dramatic, such as on contrast (34\%), defocus blue (18\%) and Gaussian noise (22\% comparing to joint-training, and 16\% comparing to the baseline). Performance on the original distribution is still almost the same, and the original error with BN is in fact slightly lower than with GN, and takes half as many epochs to converge.

We did not further experiment with BN because of two reasons:
1) The online version does not work with BN, because the problem with inaccurate batch statistics is exacerbated when training online for many (e.g. 10000) steps. 2) The baseline error for almost every corruption type is significantly higher with BN than with GN. Although unrelated to the main idea of our paper, we make the interesting note that \emph{GN significantly improves model robustness}.

\subsection{Additional Baseline: Adversarial Logit Pairing}
As discussed in the results section, \citet{hendrycks2019benchmarking} point to Adversarial Logit Pairing (ALP) \citep{kannan2018adversarial} as an effective method for improving model robustness to corruptions and perturbations, even though it was designed to defend against adversarial attacks. 
We take ALP as an additional baseline on all benchmarks based on CIFAR-10 (using GN), following the training procedure in \citet{kannan2018adversarial} and their recommended hyper-parameters.
The implementation of the adversarial attack comes from the codebase of \citet{ding2019advertorch}. 
We did not run ALP on ImageNet because the two papers we reference for this method, \citet{kannan2018adversarial} and \citet{hendrycks2019benchmarking}, did not run on ImageNet or make any claim or recommendation.

\subsection{Results on CIFAR-10-C and ImageNet-C, Level 5}
\autoref{table:full_c10c} and \autoref{table:imgnet} correspond to the bar plots in the results section. Two rows of \autoref{table:full_c10c} have been presented as \autoref{table:uda} in the main text.

\begin{figure*}
	\centering
	\includegraphics[width=0.6\textwidth]{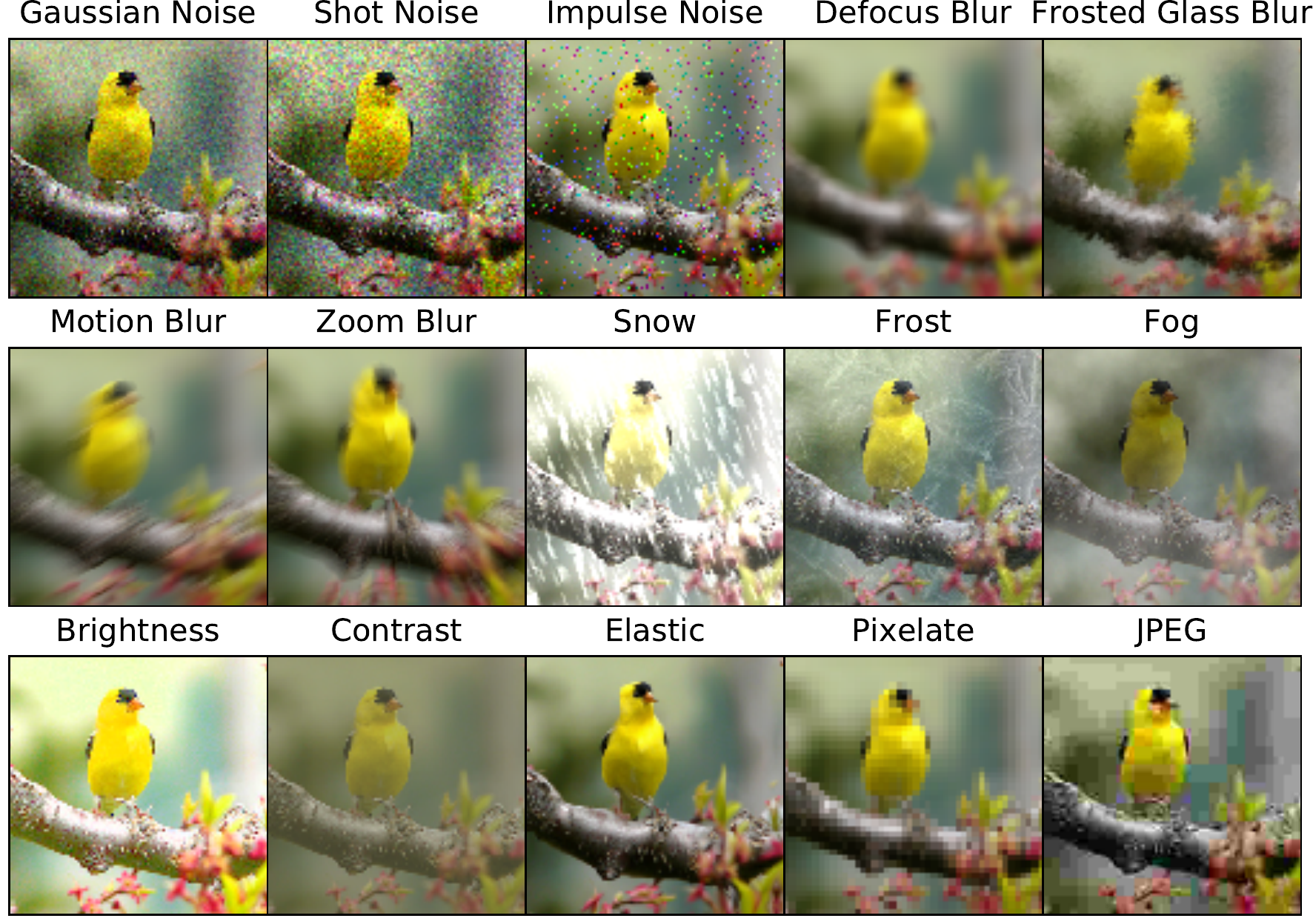}
	\caption{
		Sample images from the Common Corruptions Benchmark, taken from the original paper by \citet{hendrycks2019benchmarking}.
	}
\end{figure*}

\begin{figure*}[ht]
	\centering
	\includegraphics[width=1.0\textwidth]{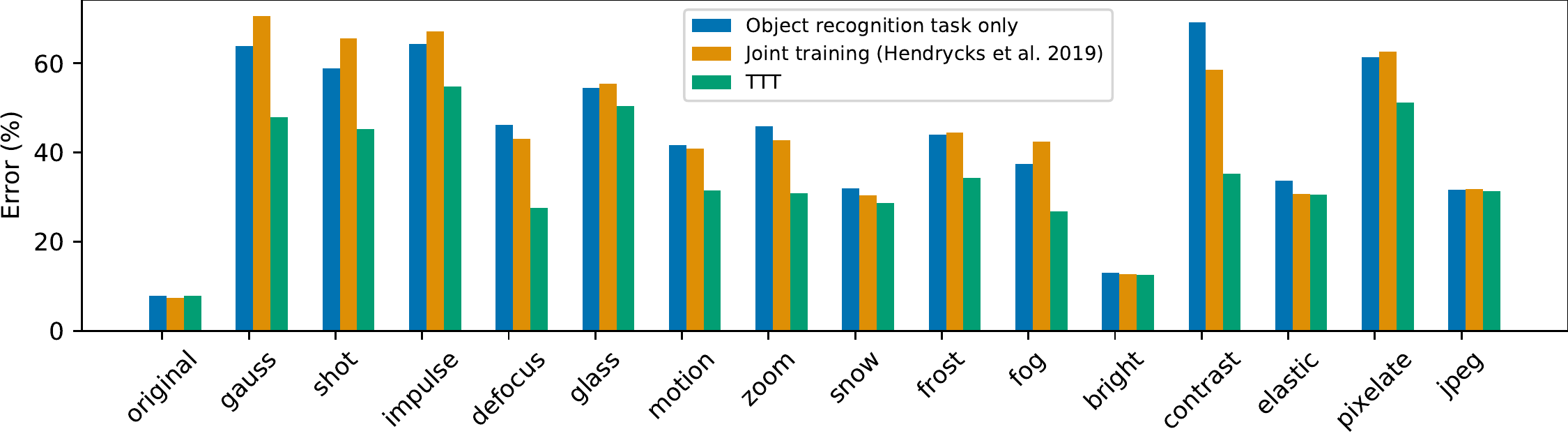}
	\caption{
		Test error (\%) on CIFAR-10-C, level 5, ResNet-26 with Batch Normalization.
	\label{figure:bn}
	}
\end{figure*}

\begin{table*}[ht]
	\footnotesize
	\begin{center}
		{
			\setlength\tabcolsep{3.5pt}
			\begin{tabular}{c|c|c|c|c|c|c|c|c|c|c|c|c|c|c|c|c}
				\hline
				& orig& gauss& shot& impul& defoc& glass& motn& zoom& snow& frost& fog& brit& contr& elast& pixel& jpeg\\
				\hline
				B & 7.9& 63.9& 58.8& 64.3& 46.3& 54.6& 41.6& 45.9& 31.9& 44.0& 37.5& 13.0& 69.2& 33.8& 61.4& 31.7\\
				\hline
				JT & 7.5& 70.7& 65.6& 67.2& 43.1& 55.4& 40.9& 42.7& 30.3& 44.5& 42.5& 12.7& 58.6& 30.7& 62.6& 31.9\\
				\hline
				TTT & 7.9& 47.9& 45.2& 54.8& 27.6& 50.4& 31.5& 30.9& 28.7& 34.3& 26.9& 12.6& 35.2& 30.6& 51.2& 31.3\\
				\hline
			\end{tabular}
		}
	\end{center}
	\vspace{-2ex}
	\caption{
		Test error (\%) on CIFAR-10-C, level 5, ResNet-26 with Batch Normalization.
	}
	\label{table:bn}
\end{table*} 

\subsection{Results on CIFAR-10-C, Levels 1-4}
The following bar plots and tables are on levels 1-4 of CIFAR-10-C.
The original distribution is the same for all levels, so are our results on the original distribution.

\subsection{Direct Comparison with \citet{hendrycks2019using}}
\label{reviewer_stuff}
The following comparison has been requested by an anonymous reviewer for our final version.
Our joint training baseline is based on \citet{hendrycks2019using}, but also incorporates some architectural changes (see below). We found these changes improved the robustness of our method, and felt that it was important to give the baseline the same benefit. Note that our joint training baseline overall performs better than Hendrycks: Compare Table S2 to Figure 3 of \citet{hendrycks2019using} (provided by the authors), our baseline has average error of 22.8\% across all corruptions and levels, while their average error is 28.6\%. 

Summary of architectural changes: 1) Group Normalization (GN) instead of Batch Normalization (BN). For completeness, the results with BN are provided in Table S1; c.f. GN results in Table S2 which significantly improves robustness, with or without self-supervision. 2) We split after the second residual group, while they split after the third residual group right before the linear layer. This consistently gives about 0.5\% - 1\% improvement. 3) We use a ResNet-26, while they use a 40-2 Wide ResNet. But our baseline still performs better than their method even though our network is 4x smaller, due to the two tricks above.

\newpage
\begin{table*}[ht]
	\footnotesize
	\begin{center}
		{
			\setlength\tabcolsep{3.5pt}
			\begin{tabular}{c|c|c|c|c|c|c|c|c|c|c|c|c|c|c|c|c}
				\hline
				& orig& gauss& shot& impul& defoc& glass& motn& zoom& snow& frost& fog& brit& contr& elast& pixel& jpeg\\
				\hline
				B & 8.9& 50.5& 47.2& 56.1& 23.7& 51.7& 24.3& 26.3& 25.6& 34.4& 28.1& 13.5& 25.0& 27.4& 55.8& 29.8\\
				\hline
				JT & 8.1& 49.4& 45.3& 53.4& 24.2& 48.5& 24.8& 26.4& 25.0& 32.5& 27.5& 12.6& 25.3& 24.0& 51.6& 28.7\\
				\hline
				TTT & 7.9& 45.6& 41.8& 50.0& 21.8& 46.1& 23.0& 23.9& 23.9& 30.0& 25.1& 12.2& 23.9& 22.6& 47.2& 27.2\\
				\hline
				TTT-Online & 8.2& 25.8& 22.6& 30.6& 14.6& 34.4& 18.3& 17.1& 20.0& 18.0& 16.9& 11.2& 15.6& 21.6& 18.1& 21.2\\
				\hline
				UDA-SS & 9.0& 28.2& 26.5& 20.8& 15.6& 43.7& 24.5& 23.8& 25.0& 24.9& 17.2& 12.7& 11.6& 22.1& 20.3& 22.6\\
				\hline
				ALP & 16.5& 22.7& 22.9& 28.3& 25.0& 25.6& 27.4& 23.1& 25.2& 27.2& 64.8& 21.7& 73.6& 23.0& 20.2& 18.9\\
				\hline
			\end{tabular}
		}
	\end{center}
	\vspace{-2ex}
	\caption{
		Test error (\%) on CIFAR-10-C, level 5, ResNet-26.
	}
\label{table:full_c10c}
\end{table*} 

\begin{table*}[ht]
	\footnotesize
	\begin{center}
		{
			\setlength\tabcolsep{3.5pt}
			\begin{tabular}{c|c|c|c|c|c|c|c|c|c|c|c|c|c|c|c|c}
				\hline
				& orig& gauss& shot& impul& defoc& glass& motn& zoom& snow& frost& fog& brit& contr& elast& pixel& jpeg\\
				\hline
				B & 68.9& 1.3& 2.0& 1.3& 7.5& 6.6& 11.8& 16.2& 15.7& 14.9& 15.3& 43.9& 9.7& 16.5& 15.3& 23.4\\
				\hline
				JT & 69.1& 2.1& 3.1& 2.1& 8.7& 6.7& 12.3& 16.0& 15.3& 15.8& 17.0& 45.3& 11.0& 18.4& 19.7& 22.9\\
				\hline
				TTT & 69.0& 3.1& 4.5& 3.5& 10.1& 6.8& 13.5& 18.5& 17.1& 17.9& 20.0& 47.0& 14.4& 20.9& 22.8& 25.3\\
				\hline
				TTT-Online & 68.8& 26.3& 28.6& 26.9& 23.7& 6.6& 28.7& 33.4& 35.6& 18.7& 47.6& 58.3& 35.3& 44.3& 47.8& 44.3\\
				\hline
			\end{tabular}
		}
	\end{center}
	\vspace{-2ex}
	\caption{
		Test accuracy (\%) on ImageNet-C, level 5, ResNet-18.
	}
	\label{table:imgnet}
\end{table*} 

\newpage
\begin{figure*}[ht]
	\centering
	\includegraphics[width=1.0\textwidth]{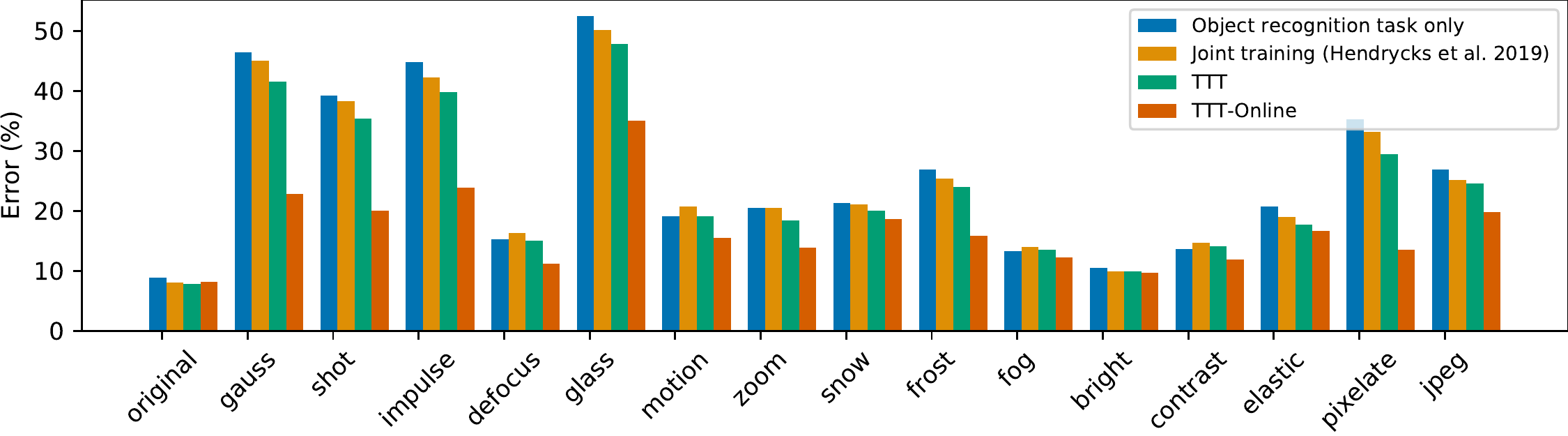}
	\vspace{-4ex}
	\caption{
		Test error (\%) on CIFAR-10-C, level 4.
		See the results section for details.
	}
	\vspace{4ex}
\end{figure*}

\begin{table*}[ht]
	\footnotesize
	\begin{center}
		{
			\setlength\tabcolsep{3.5pt}
			\begin{tabular}{c|c|c|c|c|c|c|c|c|c|c|c|c|c|c|c|c}
				\hline
				& orig& gauss& shot& impul& defoc& glass& motn& zoom& snow& frost& fog& brit& contr& elast& pixel& jpeg\\
				\hline
				B & 8.9& 46.4& 39.2& 44.8& 15.3& 52.5& 19.1& 20.5& 21.3& 26.9& 13.3& 10.5& 13.7& 20.8& 35.3& 26.9\\
				\hline
				JT & 8.1& 45.0& 38.3& 42.2& 16.4& 50.2& 20.7& 20.5& 21.1& 25.4& 14.1& 10.0& 14.7& 19.0& 33.2& 25.1\\
				\hline
				TTT & 7.9& 41.5& 35.4& 39.8& 15.0& 47.8& 19.1& 18.4& 20.1& 24.0& 13.5& 10.0& 14.1& 17.7& 29.4& 24.5\\
				\hline
				TTT-Online & 8.2& 22.9& 20.0& 23.9& 11.2& 35.1& 15.6& 13.8& 18.6& 15.9& 12.3& 9.7& 11.9& 16.7& 13.6& 19.8\\
				\hline
				ALP & 16.5& 21.3& 20.5& 24.5& 20.7& 25.9& 23.7& 21.4& 24.2& 23.9& 42.2& 17.5& 53.7& 22.1& 19.1& 18.5\\
				\hline
			\end{tabular}
		}
	\end{center}
	\vspace{-2ex}
	\caption{
		Test error (\%) on CIFAR-10-C, level 4, ResNet-26.
	} 
\end{table*} 

\begin{figure*}[ht]
	\vspace{8ex}
	\centering
	\includegraphics[width=1.0\textwidth]{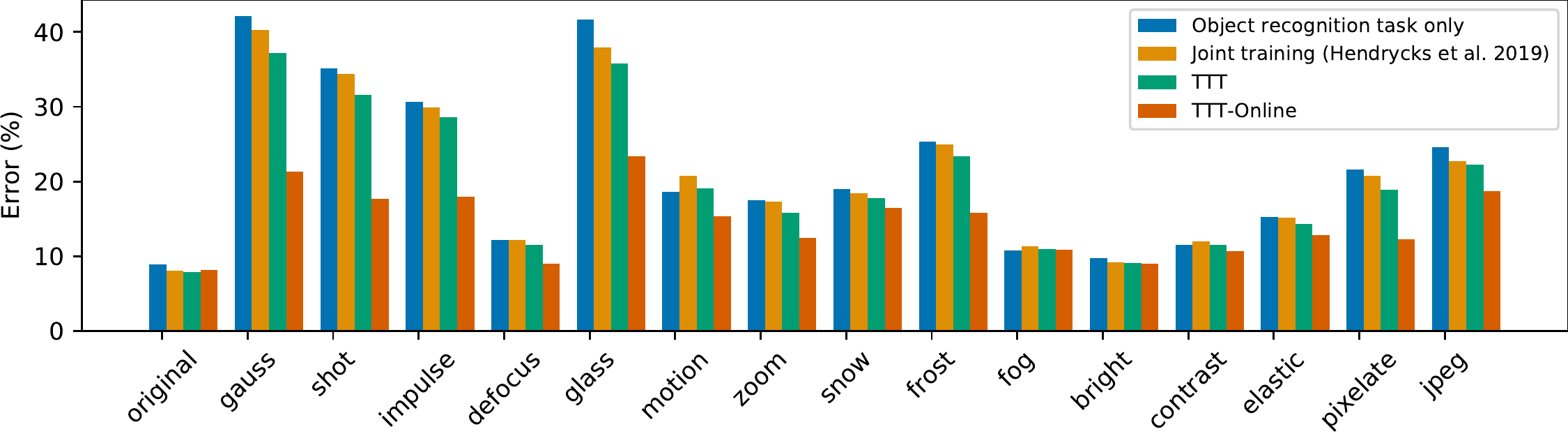}
	\vspace{-4ex}
	\caption{
		Test error (\%) on CIFAR-10-C, level 3.
		See the results section for details.
	}
\end{figure*}

\begin{table*}[ht]
	\vspace{4ex}
	\footnotesize
	\begin{center}
		{
			\setlength\tabcolsep{3.5pt}
			\begin{tabular}{c|c|c|c|c|c|c|c|c|c|c|c|c|c|c|c|c}
				\hline
				& orig& gauss& shot& impul& defoc& glass& motn& zoom& snow& frost& fog& brit& contr& elast& pixel& jpeg\\
				\hline
				B & 8.9& 42.2& 35.1& 30.7& 12.2& 41.7& 18.6& 17.5& 19.0& 25.3& 10.8& 9.7& 11.6& 15.3& 21.7& 24.6\\
				\hline
				JT & 8.1& 40.2& 34.4& 29.9& 12.2& 37.9& 20.8& 17.3& 18.4& 25.0& 11.4& 9.2& 12.0& 15.2& 20.8& 22.8\\
				\hline
				TTT & 7.9& 37.2& 31.6& 28.6& 11.5& 35.8& 19.1& 15.8& 17.8& 23.3& 11.0& 9.1& 11.6& 14.3& 18.9& 22.3\\
				\hline
				TTT-Online & 8.2& 21.3& 17.7& 17.9& 9.0& 23.4& 15.3& 12.5& 16.4& 15.8& 10.9& 9.0& 10.7& 12.8& 12.2& 18.7\\
				\hline
				ALP & 16.5& 20.0& 19.3& 20.5& 19.2& 21.2& 24.0& 20.5& 20.9& 24.2& 30.1& 16.6& 39.6& 20.9& 17.8& 18.0\\
				\hline
			\end{tabular}
		}
	\end{center}
	\caption{
		Test error (\%) on CIFAR-10-C, level 3, ResNet-26.
	} 
\end{table*}
 
\begin{figure*}[ht]
	\vspace{-1ex}
	\centering
	\includegraphics[width=1.0\textwidth]{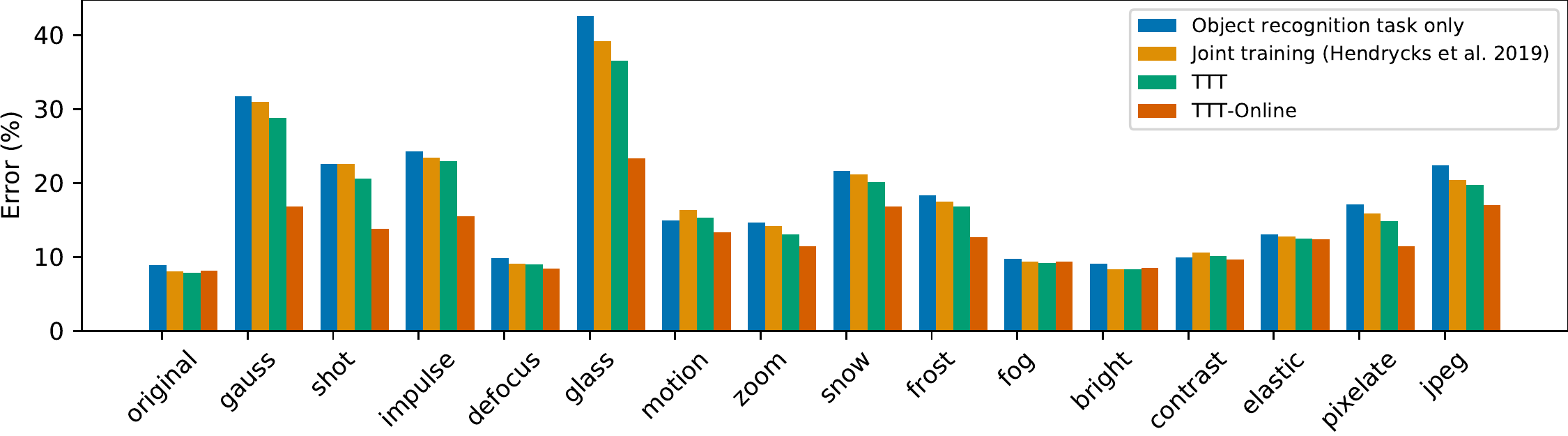}
	\vspace{-4ex}
	\caption{
		Test error (\%) on CIFAR-10-C, level 2.
		See the results section for details.
	}
	\vspace{-1ex}
\end{figure*}

\begin{table*}[ht]
	\footnotesize
	\begin{center}
		{
			\setlength\tabcolsep{3.5pt}
			\begin{tabular}{c|c|c|c|c|c|c|c|c|c|c|c|c|c|c|c|c}
				\hline
				& orig& gauss& shot& impul& defoc& glass& motn& zoom& snow& frost& fog& brit& contr& elast& pixel& jpeg\\
				\hline
				B & 8.9& 31.7& 22.6& 24.3& 9.9& 42.6& 14.9& 14.7& 21.7& 18.4& 9.8& 9.1& 10.0& 13.1& 17.1& 22.4\\
				\hline
				JT & 8.1& 31.0& 22.6& 23.4& 9.1& 39.2& 16.4& 14.2& 21.2& 17.5& 9.4& 8.3& 10.6& 12.8& 15.9& 20.5\\
				\hline
				TTT & 7.9& 28.8& 20.7& 23.0& 9.0& 36.6& 15.4& 13.1& 20.2& 16.9& 9.2& 8.3& 10.2& 12.5& 14.8& 19.7\\
				\hline
				TTT-Online & 8.2& 16.8& 13.8& 15.5& 8.5& 23.4& 13.3& 11.5& 16.8& 12.7& 9.4& 8.4& 9.7& 12.4& 11.5& 17.0\\
				\hline
				ALP & 16.5& 18.0& 17.2& 19.0& 17.8& 20.7& 21.2& 19.3& 19.0& 20.1& 22.4& 16.3& 29.2& 20.3& 17.4& 17.8\\
				\hline
			\end{tabular}
		}
	\end{center}
	\vspace{-2ex}
	\caption{
		Test error (\%) on CIFAR-10-C, level 2, ResNet-26.
	} 
\end{table*} 

\begin{figure*}[ht]
	\vspace{-1ex}
	\centering
	\includegraphics[width=1.0\textwidth]{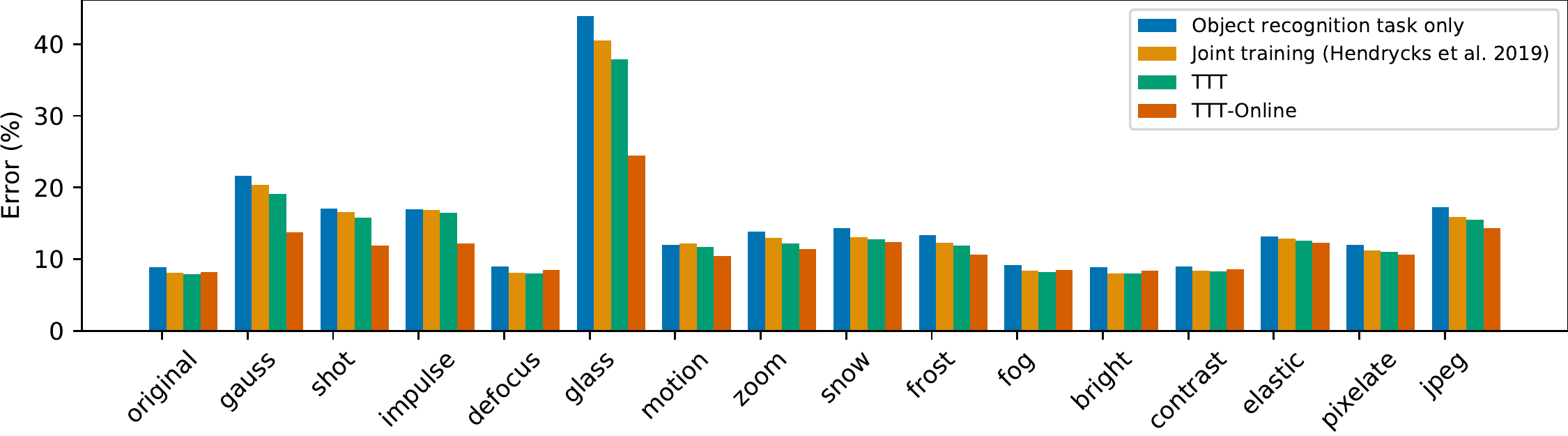}
	\vspace{-4ex}
	\caption{
		Test error (\%) on CIFAR-10-C, level 1.
		See the results section for details.
	}
	\vspace{-1ex}
\end{figure*}

\begin{table*}[ht]
	\footnotesize
	\begin{center}
		{
			\setlength\tabcolsep{3.5pt}
			\begin{tabular}{c|c|c|c|c|c|c|c|c|c|c|c|c|c|c|c|c}
				\hline
				& orig& gauss& shot& impul& defoc& glass& motn& zoom& snow& frost& fog& brit& contr& elast& pixel& jpeg\\
				\hline
				B & 8.9& 21.7& 17.1& 17.0& 9.0& 44.0& 12.1& 13.9& 14.3& 13.4& 9.2& 8.9& 9.0& 13.2& 12.0& 17.3\\
				\hline
				JT & 8.1& 20.4& 16.6& 16.9& 8.2& 40.5& 12.2& 13.0& 13.1& 12.3& 8.4& 8.1& 8.5& 12.9& 11.3& 15.9\\
				\hline
				TTT & 7.9& 19.1& 15.8& 16.5& 8.0& 37.9& 11.7& 12.2& 12.8& 11.9& 8.2& 8.0& 8.3& 12.6& 11.1& 15.5\\
				\hline
				TTT-Online & 8.2& 13.8& 11.9& 12.2& 8.5& 24.4& 10.5& 11.5& 12.4& 10.7& 8.5& 8.3& 8.6& 12.4& 10.7& 14.4\\
				\hline
				ALP & 17.0& 16.8& 17.6& 16.8& 20.9& 18.7& 19.0& 17.3& 17.5& 17.4& 16.1& 18.4& 20.4& 17.0& 17.2 & 17.5\\
				\hline
			\end{tabular}
		}
	\end{center}
	\vspace{-2ex}
	\caption{
		Test error (\%) on CIFAR-10-C, level 1, ResNet-26.
	} 
\end{table*} 

\newpage
\begin{figure*}[ht]
	\centering
	\includegraphics[width=0.9\textwidth]{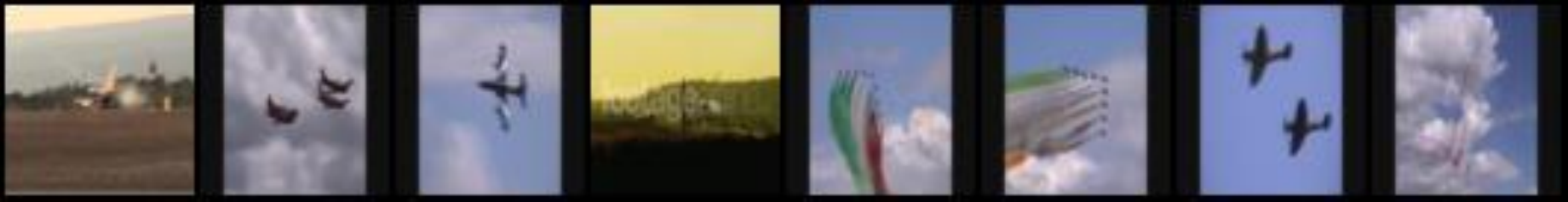}
	\includegraphics[width=0.9\textwidth]{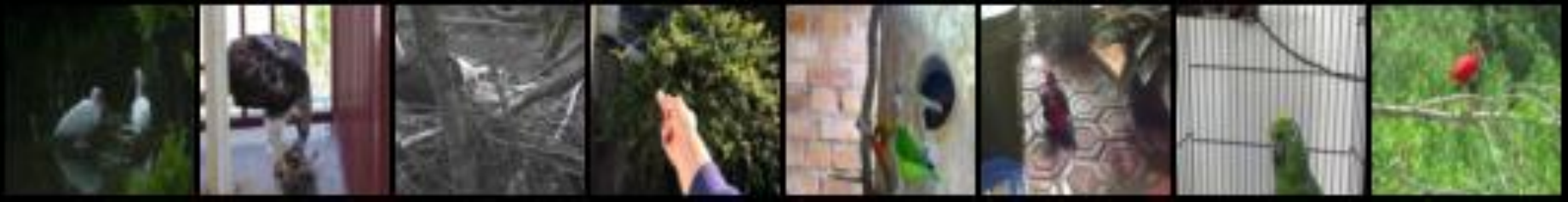}
	\includegraphics[width=0.9\textwidth]{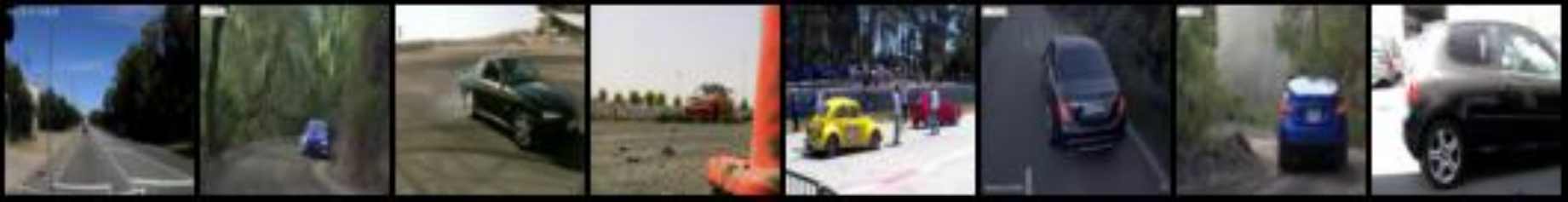}	\includegraphics[width=0.9\textwidth]{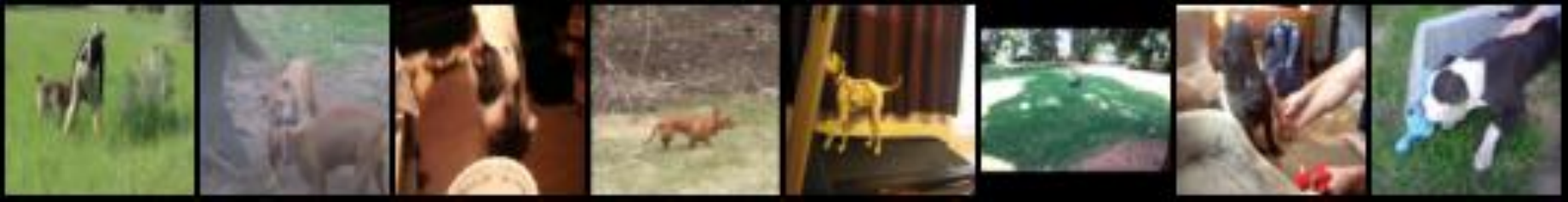}	\includegraphics[width=0.9\textwidth]{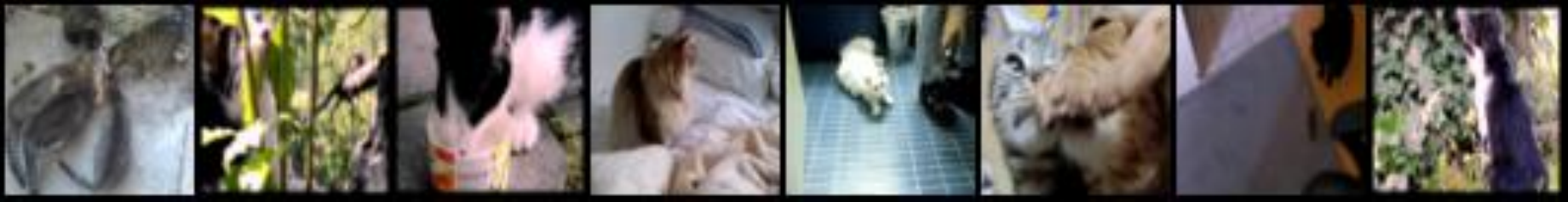}	\includegraphics[width=0.9\textwidth]{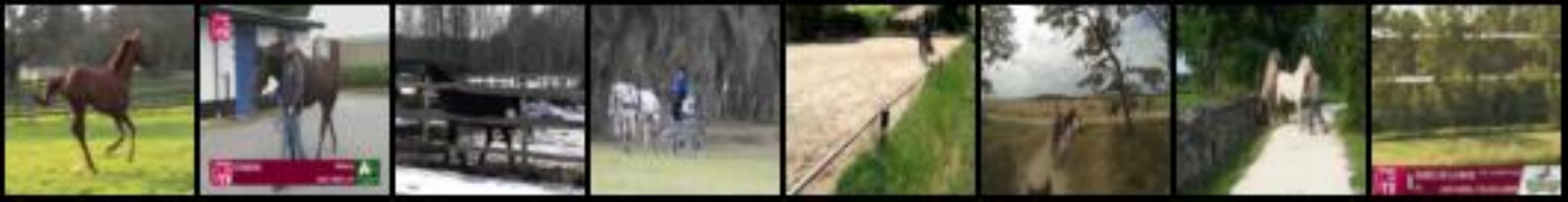}
	\includegraphics[width=0.9\textwidth]{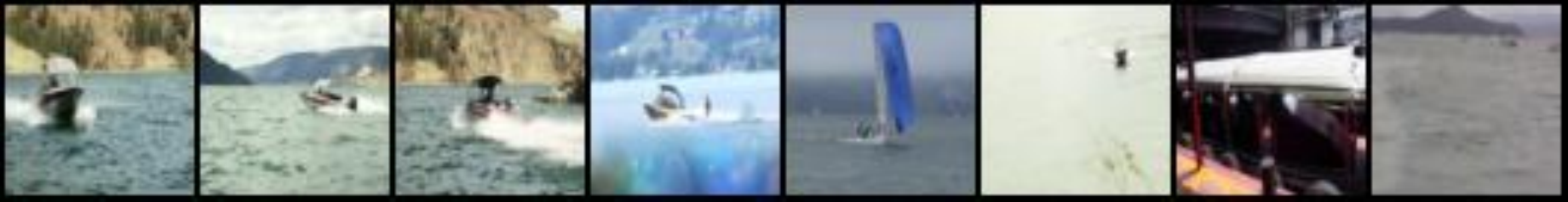}
	\caption{
		Sample Images from the VID-Robust dataset \citep{shankar2019systematic} in
		the results section adapted to CIFAR-10. 
		Each row shows eight sample images from one class.
		The seven classes shown are, in order:
		airplane, bird, car, dog, cat, horse, ship.
	}
	\label{vid_samples}
\end{figure*}
\end{document}